\documentclass[10pt,twocolumn,letterpaper]{article}

\usepackage{iccv}
\usepackage{times}
\usepackage{epsfig}
\usepackage{graphicx}
\usepackage{amsmath}
\usepackage{colortbl}
\usepackage{amssymb}
\usepackage{booktabs}
\usepackage{enumitem}
\usepackage{caption}
\usepackage{bm}
\usepackage[pagebackref=true,breaklinks=true,letterpaper=true,colorlinks,bookmarks=false]{hyperref}

\iccvfinalcopy 

\def\iccvPaperID{8084} 

\ificcvfinal\fi

\begin{document}

\definecolor{yellow}{rgb}{1,1, 0.6}
\definecolor{lightyellow}{rgb}{1,1, 0.8}
\definecolor{orange}{rgb}{1, 0.8, 0.6}
\definecolor{red}{rgb}{1, 0.6, 0.6}

\definecolor{darkyellow}{rgb}{0.8, 0.8, 0.5}
\definecolor{darkred}{rgb}{0.7, 0.3, 0.3}
\definecolor{darkgreen}{rgb}{0.3, 0.7, 0.3}
\definecolor{blue}{rgb}{0, 0, 1.0}
\definecolor{green}{rgb}{0, 1.0, 0}
\definecolor{pink}{rgb}{1, 0.4, 0.7}

\newcommand{\todo}[1]{{\color{pink}#1}}
\newcommand{\matt}[1]{{\color{blue} matt: #1}}
\newcommand{\ak}[1]{{\color{darkred} AK: #1}}
\newcommand{\hao}[1]{{\color{orange} HL: #1}}

\newcommand{\inlinepara}[1]{\noindent {\bf #1}\,\,\,}


\newcommand\estimate[1]{\hat{#1}}
\newcommand{\ray}{\mathbf{r}}
\newcommand{\raybatch}{\mathcal{R}}
\newcommand{\position}{\mathbf{x}}
\newcommand{\direction}{\mathbf{d}}
\newcommand{\col}{c}
\newcommand{\Col}{C}
\newcommand{\truecol}{C}

\newcommand{\PaperName}{\textit{Fast}NeRF\xspace} 
\newcommand{\OurNeRF}{NeRF-SH\xspace} 

\newcommand{\tf}[1]{\mathbf{#1}}

\newcommand{\Real}{\mathbb{R}}
\newcommand{\Complex}{\mathbb{C}}
\newcommand{\sphere}{\mathbb{S}^2}
\newcommand{\sglobe}{\mathbf{p}}
\renewcommand{\l}{\ell}

\definecolor{colorfirst}{rgb}{.866,.945, 0.831} 
\definecolor{colorsecond}{rgb}{1, 0.98, 0.83} 
\definecolor{colorthird}{rgb}{1, 1, 1} 

\newcommand{\cellfirst}{\cellcolor{colorfirst}}
\newcommand{\cellsecond}{\cellcolor{colorsecond}}
\newcommand{\cellthird}{\cellcolor{colorthird}}

\newcommand{\textfirst}{\colorbox{colorfirst}}
\newcommand{\textsecond}{\colorbox{colorsecond}}
\newcommand{\textthird}{\colorbox{colorthird}}

\newcommand{\fix}[1]{{\color{darkred}#1}}

\title{PlenOctrees for Real-time Rendering of Neural Radiance Fields}

\author{
Alex Yu$^1$
\qquad
Ruilong Li$^{1,2}$
\qquad
Matthew Tancik$^1$\\
Hao Li$^{1,3}$
\qquad
Ren Ng$^1$
\qquad
Angjoo Kanazawa$^1$
\vspace{0.5em}
\\
$^1$UC Berkeley \qquad
$^2$USC Institute for Creative Technologies \qquad
$^3$Pinscreen
}

\maketitle
\ificcvfinal\thispagestyle{empty}\fi

\begin{abstract}
We introduce a method to
render Neural Radiance Fields (NeRFs) in real time
using PlenOctrees, 
an octree-based 3D representation which supports view-dependent effects. Our method can render 800$\times$800 images at more than 150 FPS,  which is over 3000 times faster than conventional NeRFs.
We do so without sacrificing quality while
preserving the ability of NeRFs to perform free-viewpoint rendering of scenes with arbitrary geometry and view-dependent effects. 
Real-time performance is achieved by pre-tabulating the NeRF into a PlenOctree. In order to preserve view-dependent effects such as specularities, we  factorize the appearance via closed-form spherical basis functions.
Specifically, we show that it is possible to train NeRFs to predict a spherical harmonic representation of radiance, removing the viewing direction as an input to the neural network.
Furthermore, we show that PlenOctrees can be directly optimized to further minimize the reconstruction loss, which leads to equal or better quality compared to competing methods. 
Moreover, this octree optimization step can be used to reduce the training time, as we no longer need to wait for the NeRF training to converge fully. 
Our real-time neural rendering approach may potentially enable new applications such as 6-DOF industrial and product visualizations, as well as next generation AR/VR systems. 
PlenOctrees are amenable to in-browser rendering as well; 
please visit the project page for the interactive online demo, as well as video and code:  \url{https://alexyu.net/plenoctrees}.
\end{abstract}

\begin{figure}[ht]
    \centering
    \includegraphics[width=\linewidth]{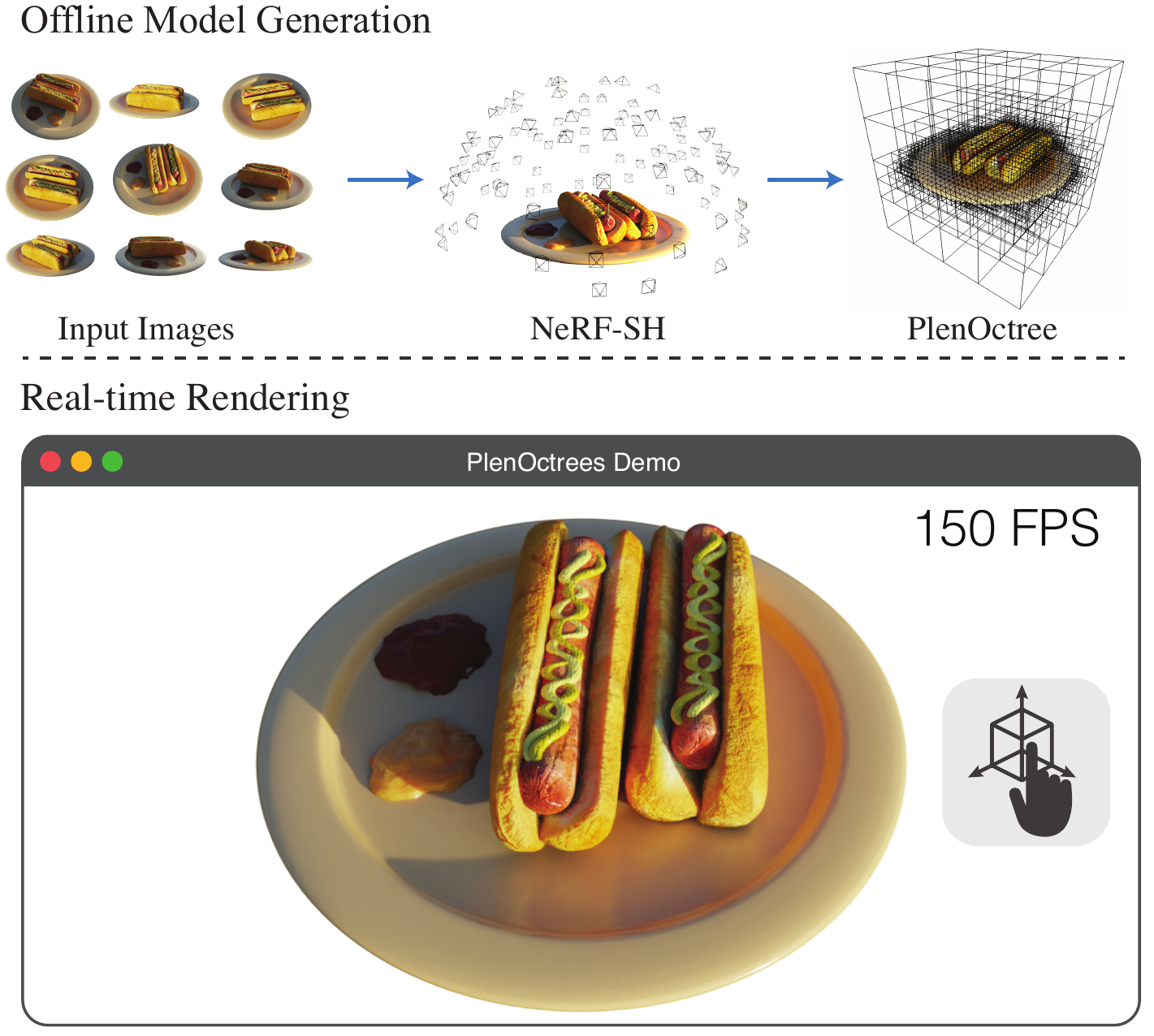}
    \caption{
    \textbf{Real-time NeRF with PlenOctrees.}
    Given a set of posed images of a scene, our method creates a 3D volumetric model that can be rendered in real-time. We propose PlenOctrees, which are octrees that can capture view-dependent dependent effects such as specularities. 
    Rendering using our approach is orders of magnitude faster than NeRF.
    }
    \label{fig:teaser}
    \vspace{-1.8em}
\end{figure}
\section{Introduction}

Despite the progress of real-time graphics, interactive 3D content with truly photorealistic scenes and objects are still time consuming and costly to produce due to the necessity of optimized 3D assets and dedicated shaders. 
Instead, many graphics applications opt for image-based solutions. E-commerce websites often use a fixed set of views to showcase their products; VR experiences often rely on 360 video recordings to avoid the costly production of real 3D scenes, and mapping services such as Google Street View stitch images into panoramic views limited to 3-DOF.

Recent advances in neural rendering,
such as neural volumes~\cite{lombardi2019neuralvol} and neural radiance fields (NeRFs)~\cite{mildenhall2020},
open a promising new avenue to model arbitrary objects and scenes in 3D 
from a set of calibrated images.
NeRFs in particular can 
faithfully render detailed scenes and appearances
with non-Lambertian effects from any view, while
simultaneously offering a high degree of compression in terms of storage.
Partly
due to these exciting properties,
of late, there has been an explosion of 
research based on NeRF.

Nevertheless,
for practical applications, runtime performance remains
a critical
limitation of NeRFs:
due to the extreme sampling requirements and
costly neural network queries, rendering a NeRF is agonizingly slow. For illustration, it takes roughly 30 seconds to render an 800x800 image from a NeRF using a high performance GPU, making it impractical for real-time interactive applications.


In this work,
we propose a method
for rendering a NeRF in real time,
achieved by
distilling the NeRF into a hierarchical 3D volumetric representation. 
Our approach preserves NeRF's ability to synthesize arbitrarily complex geometry and view-dependent effects from any viewpoint and requires no additional supervision. In fact, our method achieves and in many cases surpasses the quality of the original NeRF formulation, while providing significant acceleration.
Our model allows us to render an 800x800 image at 167.68 FPS  on a NVIDIA V100 GPU and does not rely on a deep neural network during test time. 
Moreover, our representation is amenable to modern web technologies, allowing interactive rendering in a browser on consumer laptops.

Naive NeRF rendering is slow because it requires dense sampling of the scene, where every sample requires a neural network inference. 
Because these queries depend on the viewing direction as well as the spatial position, one cannot naively cache these color values for all viewing directions.

We overcome these challenges and enable real-time rendering 
by pre-sampling the NeRF into a tabulated view-dependent volume which we refer to as a PlenOctree, named after the plenoptic functions of Adelsen and Bergen~\cite{adelson1991plenoptic}.  
Specifically, we use a sparse voxel-based octree where every leaf of the tree stores the appearance and density values required to model the radiance at a point in the volume.
In order to account for non-Lambertian materials that exhibit view-dependent effects,
we propose to represent the RGB values at a location with spherical harmonics (SH), a standard basis for functions defined on the surface of the sphere. The spherical harmonics can be evaluated at arbitrary query viewing directions to recover the view dependent color.  

\begin{figure*}[t]
    \centering
    \includegraphics[width=\linewidth]{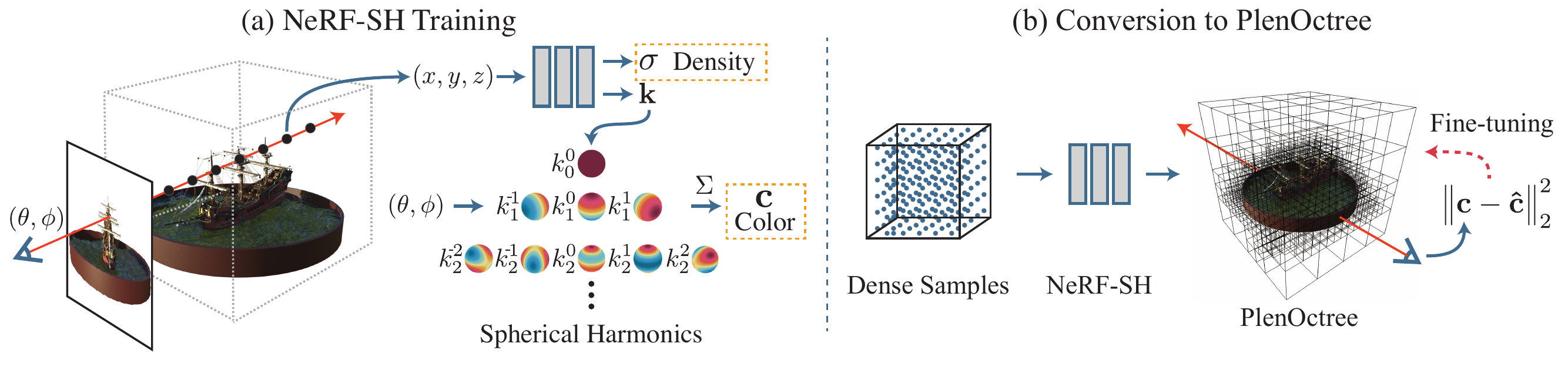}
    \caption{\textbf{Method Overview.}
    We propose a method to quickly render NeRFs by training a modified NeRF model (\OurNeRF) and converting it into a PlenOctree, an octree that  captures view-dependent effects.
    a) The \OurNeRF model uses the same optimization procedure and volume rendering method presented in NeRF~\cite{mildenhall2020}. However, instead of predicting the RGB color $\col$ directly, the network predicts spherical harmonic coefficients $\tf k$. The color $\col$ is calculated by summing the weighted spherical harmonic bases evaluated at the corresponding ray direction $(\theta,\phi)$. The spherical harmonics enable the representation to model view-dependent appearance. The values in the orange boxes are used for volume rendering. b) To build a PlenOctree, we densely sample the \OurNeRF model in the volume around the target object and tabulate the density and SH coefficients.     We can further optimize the PlenOctree directly with the training images to improve its quality.
    }
    \vspace{-1em}
    \label{fig:pipeline}
\end{figure*}

Although one could convert an existing NeRF into such a representations via projection onto the SH basis functions, we show that we can in fact modify 
a NeRF network to predict 
appearances explicitly in terms of spherical harmonics. Specifically, we train a network that produces coefficients for the SH functions instead of raw RGB values,
so that the predicted values can later be directly stored within the leaves of the PlenOctree. 
 We also introduce a sparsity prior during NeRF training to improve the memory efficiency of our octrees, consequently allowing us to render higher quality images. Furthermore, once the structure is created, the values stored in PlenOctree can be optimized because the rendering procedure remains differentiable. 
 This enables the PlenOctree to obtain similar or better image quality compared to NeRF.
 Our pipeline is illustrated in Fig.~\ref{fig:pipeline}.
 
 Additionally, we demonstrate how our proposed pipeline can be used to accelerate NeRF model training, making our solution more practical to train than the original NeRF approach. Specifically, we can stop training the NeRF model early to convert it into a PlenOctree, which can then be trained significantly faster as it no longer involves any neural networks.

Our experiments demonstrate that our approach can accelerate NeRF-based rendering by 5 orders of magnitude without loss in image quality. We compare our approach on standard benchmarks with scenes and objects captured from $360^{\circ}$ views, and demonstrate state-of-the-art level performance for image quality and rendering speed. 

Our interactive viewer can enable
operations such as object insertion, visualizing
radiance distributions,
decomposing the SH components,
and slicing the scene.
We hope that these real-time operations can be useful to the community
for visualizing and debugging NeRF-based representations.

To summarize,
we make the following contributions:
\begin{itemize}
    \item The first method that achieves real-time rendering of NeRFs with similar or improved quality. 

    \item NeRF-SH: a modified NeRF that is trained to output appearance in terms of spherical basis functions. 
    
    \item PlenOctree, 
    a data structure derived from NeRFs which enables highly efficient view-dependent rendering of complex scenes.
    \item Accelerated NeRF training method using an early training termination, followed by a direct fine-tuning process on PlenOctree values.
\end{itemize}

\section{Related Work}

\noindent\textbf{Novel View Synthesis.}
The task of synthesizing novel views of a scene given a set of photographs is a well-studied problem with various approaches. All methods predict an underlying geometric or image-based 3D representation that allows rendering from novel viewpoints.
Mesh based methods represent the scene with surfaces, and have been used to model Lambertian (diffuse)~\cite{waechter2014let} and non-Lambertian scenes~\cite{wood2000surface,debevec1996modeling,buehler2001unstructured}.

Mesh based representations are compact and easy to render; however, optimizing a mesh to fit a complex scene of arbitrary topology is challenging. Image-based rendering methods~\cite{levoy1996light,shade1998layered,wood2000surface},
on the other hand, enable easy capture as well as photo-realistic and fast rendering, however are often bounded in the viewing angle and do not allow easy editing of the underlying scene.

Volume rendering is a classical technique with a long history of research in the graphics community~\cite{drebin1988volume}.
Volume-based representations such as voxel grids~\cite{seitz1999photorealistic, kutulakos2000theory,lombardi2019neuralvol,kar2017learning,tulsiani2017multi,sitzmann2019deepvoxels} and multi-plane images (MPIs)~\cite{szeliski1998stereo,penner2017soft,zhou2018stereo,srinivasan2019pushing,mildenhall2019local} are a popular alternative to mesh representations 
due to their topology-free nature: gradient-based optimization is therefore straightforward, while rendering can still be real-time. 
However, such naive volumetric representations are often memory bound, limiting the maximum resolution that can be captured. 
Volumetric octrees are a popular approach for reducing memory and compute in such cases. We refer the reader to this survey~\cite{knoll2006survey} for a historical perspective on octree volume rendering. Octrees have been used in recent work to decrease the memory requirements during training for other 3D tasks~\cite{Riegler2017OctNet,hane2017hierarchical,tatarchenko2017octree,wang2017cnn}.
Concurrent with this work, NeX~\cite{wizadwongsa2021nex} extends MPIs to encode spherical basis functions that enable view-dependent rendering effects in real-time. Also concurrently, Lombardi et al.~\cite{lombardi2021mixture} propose to model data using geometric primitives, and~\cite{hedman2021snerg,garbin2021fastnerf,reiser2021kilonerf} also distill NeRFs to enable real-time rendering.

 \vspace{.1em}
 \noindent\textbf{Coordinate-Based Neural Networks.}
 Recently, coordinate-based neural networks have emerged as a popular alternative to explicit volumetric representations, as they are not limited to a fixed voxel representation. These methods train a multilayer perceptron (MLP) whose input is a coordinate and output is some property of space corresponding to that location. These networks have been used to predict occupancy~\cite{OccupancyNetworks,chen_cvpr19,peng2020conv,saito2019pifu,niemeyer2020dvr,li2020monocular}, signed distance fields~\cite{park2019deepsdf, gropp2020implicit,disn,yariv2020multiview}, and radiance~\cite{mildenhall2020}.
 Coordinate-based neural networks have been used for view synthesis in Scene Representation Networks~\cite{sitzmann2019scene}, NeRFs~\cite{mildenhall2020}, and many NeRF extensions~\cite{martinbrualla2020nerfw,park2020deformable,schwarz2020graf,srinivasan2020nerv}. These networks represent a continuous function that can be sampled at arbitrarily fine resolutions without increasing the memory footprint. Unfortunately, this compactness is achieved at the expense of computational efficiency as each sample must be processed by a neural network. As a result, these representations are often slow and impractical for real-time rendering.
 
\vspace{0.2em}
\inlinepara{NeRF Accelerations.}
While NeRFs are able to produce high quality results, their computationally expensive rendering leads to slow training and inference. One way to speed up the process of fitting a NeRF to a new scene is to incorporate priors learned from a dataset of similar scenes. This can be accomplished by conditioning on predicted images features~\cite{trevithick2020grf,yu2020pixelnerf,wang2021ibrnet} or meta-learning~\cite{tancik2020meta}. To improve inference speed, Neural Sparse Voxel Fields (NSVF)~\cite{nsvf} learns a sparse voxel grid of features that are input into a NeRF like model. The sparse voxel grid allows the renderer to skip over empty regions when tracing a ray which improves the render time $\sim$10x. Decomposed Radiance Fields~\cite{rebain2020derf} spatially decomposes a scene into multiple smaller networks. AutoInt~\cite{lindell2020autoint} modifies the architecture of the NeRF so that inference requires fewer samples but produces lower quality results. None of these approaches achieve real-time. The concurrent work DoNeRF adds a depth classifier to NeRF in order to drastically improve the efficiency of sampling, but requires ground-truth depth for training. Although not based on NeRF, recently Takikawa \etal~\cite{takikawa2021nglod} propose a method to accelerate neural SDF rendering with an octree.
Note that this work does not model appearance properties.
In contrast, we employ a volumetric representation that can capture photorealistic view-dependent appearances while achieving even higher framerates.

\section{Preliminaries}
\label{section:prelim}

\subsection{Neural Radiance Fields}
Neural radiance fields (NeRF)~\cite{mildenhall2020} are 3D representations that can be rendered from arbitrary novel viewpoints while capturing continuous geometry and view-dependent appearance. The radiance field is encoded into the weights of a multilayer perceptron (MLP) that can be queried at a position $\position = (x, y, z)$ from a viewing direction $\direction = (\theta, \phi)$ to recover the corresponding density $\sigma$ and color $\col = (r, g, b)$. A pixel’s predicted color $\Col(\ray)$ is computed by casting a ray, $\ray$, into the volume and accumulating the color based on density along the ray. NeRF estimates the accumulated color by taking $N$ point samples along the ray to perform volume rendering: 
\begin{align}
    &\estimate{\Col}(\ray) = \sum_{i=0}^{N-1} T_i \big(1 - \exp( -\sigma_{i} \delta_{i})\big) \, \col_{i} \, , \label{eqn:quad}\\
    &\text{where} \quad T_i = \exp\left(
        -\sum_{j=0}^{i-1} \sigma_j \delta_j
    \right) 
\end{align}
Where $\delta_i$ are the distances between point samples. To train the NeRF network, the predicted colors $\estimate{\Col}$ for a batch of rays $\raybatch$ corresponding to pixels in the training images are optimized using Adam~\cite{kingma2014adam} to match the target pixel colors:

\begin{equation}
    \mathcal{L}_{\text{RGB}} = \sum_{\ray \in \mathcal{\raybatch}} \big\lVert{\truecol(\ray) - \estimate{\Col}(\ray)}\big\rVert^2_2
    \label{eqn:nerfloss}
\end{equation}

To better represent high frequency details in the scene the inputs are positionally encoded and two stages of sampling are performed, one coarse and one fine. We refer the interested reader to the NeRF paper~\cite{mildenhall2020} paper for details. 

\paragraph{Limitations.}
One notable consequence of this architecture is that each sample along the ray must be fed to the MLP to obtain the corresponding $\sigma_i$ and $\col_i$. A total of 192 samples were taken for each ray in the examples presented in NeRF. This is inefficient as most samples are sampling free space which do not contribute to the integrated color. To render a single target image at $800 \times 800$ resolution, the network must be run on over 100 million inputs. Therefore it takes about 30 seconds to render a single frame using a NVIDIA V100 GPU, making it impractical for real-time applications. Our use of a sparse voxel octree avoids excess compute in regions without content. Additionally we precompute the values for each voxel so that network queries are not performed during inference.


\section{Method}
We propose a pipeline that enables real-time rendering of NeRFs.
Given a trained NeRF, we can convert it into a
PlenOctree, an efficient data structure that is able to represent non-Lambertian effects in a scene.
Specifically, it is an octree which stores
spherical harmonics (SH) coefficients at the leaves, encoding view-dependent radiance.

To make the conversion to PlenOctree more straightforward,
we also propose
\textit{\OurNeRF},
a variant of the NeRF network which directly outputs the SH coefficients,
thus eliminating the need for a view-direction input to the network.
With this change,
the conversion can then be performed by evaluating on a uniform grid
followed by thresholding.
We fine-tune the octree on the training images 
to further improve image quality,
Please see Fig.~\ref{fig:pipeline} for a graphical illustration of our pipeline.

The conversion process leverages the continuous nature of NeRF to dynamically 
obtain the spatial structure of the octree.
We show that even with a partially trained NeRF, 
our PlenOctree is capable of producing results competitive with the fully trained NeRF.
 
\subsection{NeRF-SH: NeRF with Spherical Harmonics}
SHs have been a popular low-dimensional representation for spherical functions and have been used to model 
Lambertian surfaces~\cite{ramamoorthi2001relationship,basri2003lambertian} or even glossy surfaces~\cite{sloan2002precomputed}. 
Here we explore its use in a volumetric context. 
Specifically, we adapt the NeRF network $f$ to 
output spherical harmonics coefficients $\tf k$ , rather than RGB values.
\begin{equation}
    f(\tf x) = (\tf k , \sigma) \hspace{1em} \text{where} \hspace{1em}
    \tf k = (k_{\l}^m)_{\l:\, 0 \le \l \le \l_{\max}}^{m:\, -\l \le m \le \l}
\end{equation}
Each $k_{\l}^m \in \Real^3$ is a set of 3 coefficients corresponding to the RGB components.
In this setup, the view-dependent color $\tf c$ at a point $\position$ may be determined by querying the SH functions $Y_{\l}^m: \sphere \mapsto \Real$ at the desired viewing angle $\direction$:
\begin{equation}
    \col (\direction; \tf k) = 
    S\left(\sum_{\l=0}^{\l_{\max}}\sum_{m=-\l}^{\l} k_{\l}^m Y_{\l}^m(\direction) \right)
\end{equation}
Where $S: x \mapsto (1 + \exp(-x))^{-1}$ is the sigmoid function for normalizing the colors.
In other words, we factorize the view-dependent appearance with the SH basis, eliminating the view-direction input to the network and removing the need to sample view directions at conversion time. 
Please see the appendix for more technical discussion of SHs.
With a single evaluation of the network, we can now 
efficiently query colors from arbitrary viewing angles at inference time.
In Fig.~\ref{fig:early_stop},
one can see that NeRF-SH
training speed is similar to, but slightly faster than, NeRF (by about 10\%).

Note that we can also project a trained NeRF to SHs directly at each point by 
sampling NeRF at random directions and multiplying by the SH component values to form 
Monte Carlo estimates of the inner products. However, this sampling process takes several hours to achieve reasonable quality
and imposes a quality loss of about 2 dB.\footnote{With 10000 view-direction samples per point, taking about 2 hours,
the PSNR is 29.21 vs.~31.02 for our main method prior to optimization.
}
Nevertheless, this alternative approach offers a pathway to 
convert existing NeRFs into PlenOctrees.

Other than SHs, we also experiment with Spherical Gaussians (SG)~\cite{fisher1953dispersion},
a learnable spherical basis which have been used to represent all-frequency lighting~\cite{tsai2006all, sloan2002precomputed, li2020inverse}.
We find that SHs perform better in our use case and provide an ablation in the appendix.

\vspace{0.2em}
\noindent\textbf{Sparsity prior.}
Without any regularization, the model is free to generate arbitrary geometry in unobserved regions.
While this does not directly worsen image quality,
it would adversely impact our conversion process as 
the extra geometry occupies significant voxel space.

\begin{figure}
     \centering
     \includegraphics[width=\linewidth]{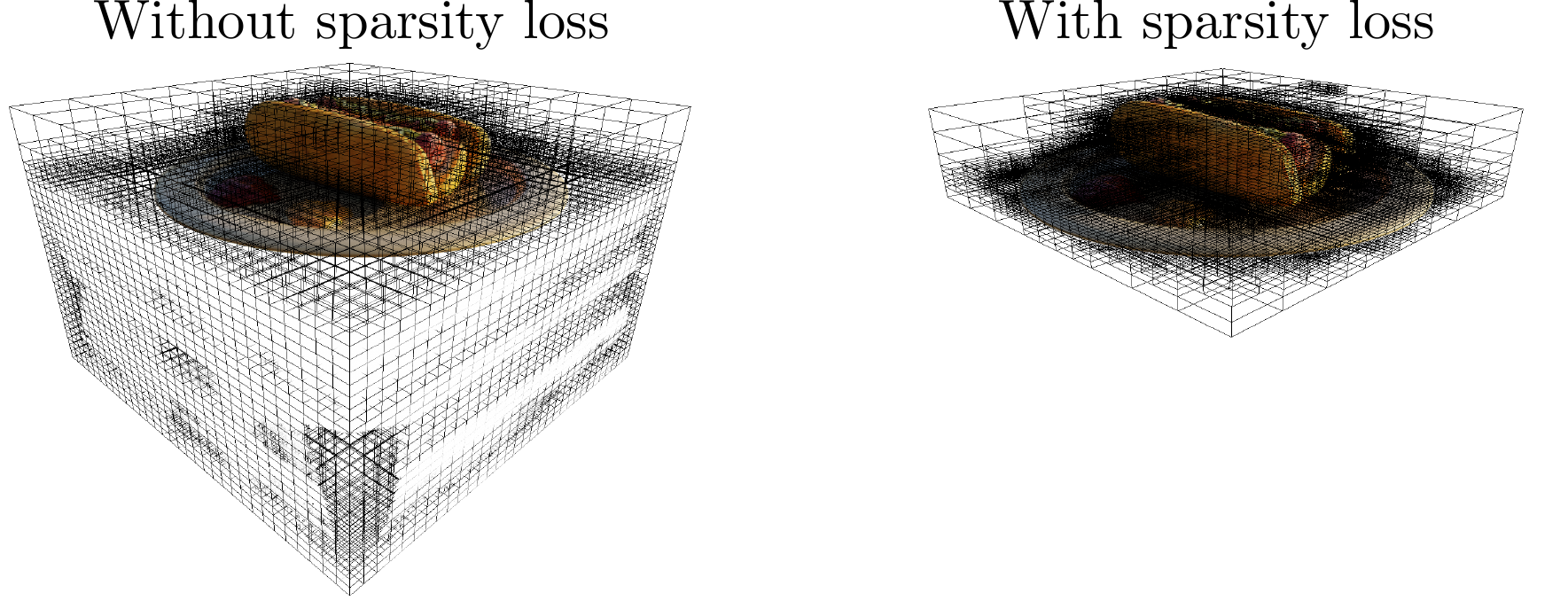}
     \caption{\textbf{Sparsity loss and conversion robustness}. When trained without the sparsity loss, NeRF
      can often converge to a solution where unobserved portions or
     the background are solid.
     This degrades the spatial resolution of our octree-based representation. 
     }
     \label{fig:sparsity_loss}
 \end{figure}

To solve this problem, we introduce an additional sparsity prior during NeRF training.
Intuitively, this prior encourages NeRF to choose empty space when both space and solid colors are possible solutions. Formally,
\begin{equation}
    \mathcal{L}_{\text{sparsity}} =
    \frac1{K} \sum_{k=1}^K \left\lvert 1 - \exp(-\lambda \sigma_k) \right\rvert
\end{equation}
Here, $\{\sigma_k\}_{k=1}^K$ are the evaluated density values at
$K$ uniformly random points within the bounding box, and
$\lambda$ is a hyperparameter.
The final training loss is then $\beta_{\text{sparsity}}\mathcal{L}_{\text{sparsity}} + \mathcal{L}_{\text{RGB}}$, where $\beta_{\text{sparsity}}$ is a hyperparameter.
Fig.~\ref{fig:sparsity_loss} illustrates the effect of the prior. 

\begin{table}[]
    \centering
    \resizebox{\linewidth}{!}{
    \begin{tabular}{@{}l c ccc c c}
\multicolumn{7}{c}{Synthetic NeRF Dataset \hspace{1em}\textfirst{best} \textsecond{second-best}}
\vspace{0.1em}\\
\toprule
 &  & PSNR $\uparrow$ & SSIM $\uparrow$ & LPIPS $\downarrow$ &  & FPS $\uparrow$ \\
\cmidrule{1-1} \cmidrule{3-5} \cmidrule{7-7}
NeRF (original) &  & 31.01 & 0.947 & 0.081 &  & 0.023 \\
NeRF &  & \cellthird 31.69 & \cellthird 0.953 & 0.068 &  & 0.045 \\
SRN &  & 22.26 & {0.846} & 0.170 &  & \cellthird 0.909 \\
Neural Volumes &  & {26.05} & {0.893} & {0.160} &  & \cellsecond 3.330 \\
NSVF &  & \cellfirst 31.75 & \cellsecond 0.953 & \cellfirst 0.047 &  & {0.815} \\
AutoInt (8 sections) &  & 25.55 & 0.911 & 0.170 &  & 0.380 \\
\cmidrule{1-1} \cmidrule{3-5} \cmidrule{7-7}
\OurNeRF &  & 31.57 & 0.952 & \cellthird 0.063 &  & 0.051 \\
PlenOctree from \OurNeRF &  & 31.02 & 0.951 & 0.066 &  & 167.68 \\
PlenOctree after fine-tuning &  & \cellsecond 31.71 & \cellfirst 0.958 & \cellsecond 0.053 &  & \cellfirst 167.68 \\
\bottomrule
\end{tabular}
    }
    \vspace{0.1em}
    \caption{\textbf{Quantitative results on the NeRF-synthetic test scenes.} Our approach is significantly faster than all existing methods during inference while performing on par with \textit{NSVF}, the current state-of-the-art method for image quality.
    We note that \textit{\OurNeRF},
    the modified NeRF model that is trained to output SH, performs similarly to the baseline \textit{NeRF} model. The octree conversion of \textit{\OurNeRF} to
    \textit{PlenOctree w/o fine-tuning} negatively impacts the image quality metrics. This is remedied with the additional fine-tuning step.}
    \label{tab:synthetic_results}
\end{table}

\begin{table}[]
    \centering
    \resizebox{\linewidth}{!}{
    \begin{tabular}{@{}l c ccc c c}
\multicolumn{7}{c}{Tanks and Temples Dataset \hspace{1em}\textfirst{best} \textsecond{second-best}}
\vspace{0.1em}\\
\toprule
 &  & PSNR $\uparrow$ & SSIM $\uparrow$ & LPIPS $\downarrow$ &  & FPS $\uparrow$ \\
\cmidrule{1-1} \cmidrule{3-5} \cmidrule{7-7}
NeRF (original) &  & 25.78 & 0.864 & 0.198 &  & 0.007 \\
NeRF &  & \cellthird 27.94 & \cellsecond 0.904 & 0.168 &  & 0.013 \\
SRN &  & 24.10 & 0.847 & 0.251 &  & \cellthird 0.250 \\
Neural Volumes &  & 23.70 & 0.834 & 0.260 &  & \cellsecond 1.000 \\
NSVF &  & \cellfirst 28.40 & 0.900 & \cellsecond 0.153 &  & 0.163 \\
\cmidrule{1-1} \cmidrule{3-5} \cmidrule{7-7}
\OurNeRF &  & 27.82 & \cellthird 0.902 & \cellthird 0.167 &  & 0.015 \\
PlenOctree from \OurNeRF &  & 27.34 & 0.897 & 0.170 &  & 42.22 \\
PlenOctree after fine-tuning &  & \cellsecond 27.99 & \cellfirst 0.917 & \cellfirst 0.131 &  & \cellfirst 42.22 \\
\bottomrule
\end{tabular}

    }
    \vspace{0.1em}
    \caption{\textbf{Quantitative results on the Tanks and Temples test scenes.}
    We find that our fine-tuned \textit{PlenOctree} model is significantly faster than existing methods while performing on par in terms of image metrics.
    Note that the images here are 1920$\times$1080
    compared to 800$\times$800 in the
    synthetic dataset.}
    \label{tab:tt_results}
\end{table}

\subsection{PlenOctree: Octree-based Radiance Fields}
\label{subsection:octree}
Once we have trained a \OurNeRF model,
we can convert it into a sparse octree representation for real time rendering.
A PlenOctree stores density and SH coefficients modelling view-dependent appearance at each leaf.
We describe the conversion and rendering processes below. 

\vspace{0.2em}
\noindent\textbf{Rendering.}
To render the PlenOctree,
for each ray, we 
first determine ray-voxel intersections in the octree structure.
This produces a sequence of segments between voxel boundaries 
with constant density and color, of lengths $\{\delta_i\}_{i=1}^{N}$.
NeRF's volume rendering model~\eqref{eqn:quad} is then applied to 
assign a color
to the ray. 
This approach allows for skipping large voxels in one step while also not missing small voxels.

At test-time, we further accelerate this rendering process by applying 
early-stopping when the ray has accumulated transmittance $T_i$ less
than $\gamma = 0.01$.

\vspace{0.3em}
\noindent\textbf{Conversion from \OurNeRF.}
The conversion process can be divided into three steps.
At a high level, we evaluate the network on a grid, retaining only density values,
then filter the voxels via thresholding.
Finally we sample random points within each remaining voxel 
and average them to obtain SH coefficients to store in the octree leaves.
More details are given below:

\textit{Evaluation}. 
 We first evaluate the \OurNeRF model to obtain $\sigma$ values
on a uniformly spaced 3D grid. The grid is automatically scaled to tightly fit the scene content.\footnote{
    By pre-evaluating $\sigma$ on a larger grid and
    finding the bounding box of all points with $\sigma \ge \tau_a$.
}

\begin{figure}[]
    \centering
    \includegraphics[width=\linewidth]{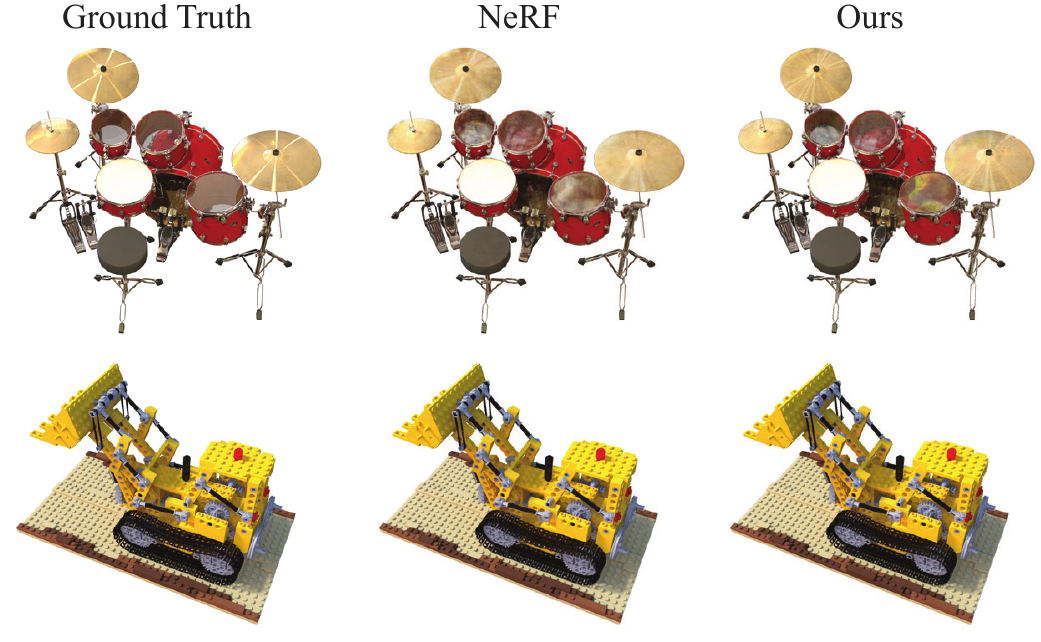} 
    \caption{\textbf{NeRF-synthetic qualitative results.} Randomly sampled qualitative comparisons on a reimplementation of NeRF and our proposed method. We are unable to find any significant image quality difference between the two methods. Despite this, our method can render these examples images more than 3500x faster.}
    \label{fig:qual_comparison}
    \vspace{-0.5em}
\end{figure}

  \textit{Filtering}. 
Next, we filter this grid to obtain a sparse set of voxels
centered at the grid points sufficient for representing the scene.
Specifically,
we render alpha maps for all the training views using this voxel grid,
keeping track of the maximum ray weight \(1 - \exp(-\sigma_i \delta_i)\) at each voxel.
We then eliminate the voxels whose weights are lower than a threshold $\tau_w$.
The octree is constructed 
to contain the remaining voxels as leaves at the deepest level while being empty elsewhere.
Compared to naively thresholding by $\sigma$ at each point,
this method eliminates non-visible voxels.

    \textit{Sampling}. 
Finally, we sample a set of $256$ random points in each remaining voxel and
set the associated leaf of the octree to the mean 
of these values to reduce aliasing.
Each leaf now contains the
density $\sigma$ and a vector of spherical harmonics coefficients for each of the RGB color channels. 

This full extraction process takes about 15 minutes.\footnote{
Note that sampling $8$ points instead of $256$ allows for extraction in about $1.5$ minutes,
with minimal loss in quality.
}

\subsection{PlenOctree Optimization}
\label{subsection:finetune}
Since this volume rendering process is fully differentiable with respect to the tree values,
we can directly fine-tune the resulting octree
on the original training images using the NeRF loss \eqref{eqn:nerfloss} with SGD 
 in order to improve the image quality.
Note that the tree structure is fixed to that obtained from NeRF in this process.
PlenOctree optimization operates at about $3$ million rays per second,
compared to about $9000$ for NeRF training, allowing us to 
optimize for many epochs in a
relatively short time.
The analytic derivatives for this process are implemented in custom CUDA kernels.
We defer technical details to the appendix.

\begin{figure}[t]
    \centering
    \includegraphics[width=\linewidth]{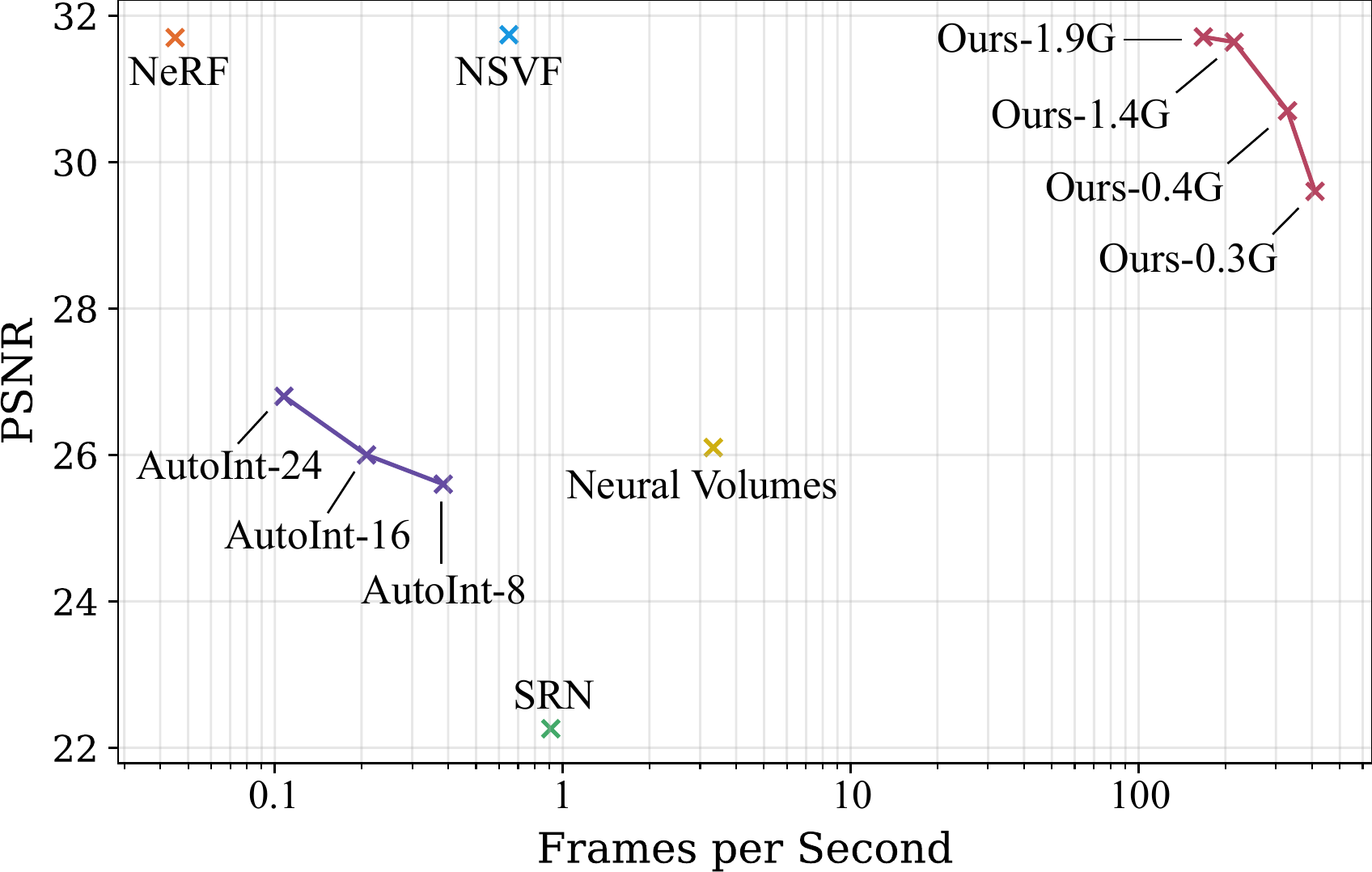}
    \caption{\textbf{Quality vs. speed comparison of various methods.} A comparison of methods on the NeRF-synthetic dataset where higher PSNR and lower FPS (top right) is most desirable. We include four variants of the PlenOctree model that tune parts of the conversion process to trade off accuracy for speed. 
    Please see Table~\ref{tab:ablation_orf}
    (adjacent) for descriptions of these variants.
    Note that the time axis is logarithmic.}
    \label{fig:time_tradeoff}
\end{figure}

\begin{table}[t]
    \centering
    \resizebox{\linewidth}{!}{
    \begin{tabular}{@{}l c c c ccc}
\vspace{0.1em}\\
\toprule
 &  & Model Description &  & GB $\downarrow$ & PSNR $\uparrow$ & FPS $\uparrow$ \\
\cmidrule{1-1} \cmidrule{3-3} \cmidrule{5-7}
Ours-1.9G &  & Complete Model as in Table 1 &  & 1.93 & 31.71 & 168 \\
Ours-1.4G &  & Higher Threshold $\tau_w=0.01$ &  & 1.36 & 31.64 & 215 \\
Ours-0.4G &  & w/o Auto Bbox Scaling &  & 0.44 & 30.70 & 329 \\
Ours-0.3G &  & Grid Size 256 &  & 0.30 & 29.60 & 410 \\
\bottomrule
\end{tabular}
    }
    \vspace{0.1em}
    \caption{\textbf{PlenOctree conversion ablations on NeRF-synthetic.} Average metrics across the NeRF-synthetic scenes for 
    several different methods to construct PlenOctrees are shown.
\textit{Ours-1.9G:} This is the high-quality model we reported in Table~\ref{tab:synthetic_results}.
\textit{Ours-1.4G:} This is a variant with a higher weight threshold, therefore aggressively sparsifying the tree.
\textit{Ours-0.4G:} Here, we remove the auto bounding box scaling step
and instead use fixed large bounding boxes, limiting resolution.
\textit{Ours-0.3G:} A version using a $256^3$ grid instead of $512^3$.
    }
    \label{tab:ablation_orf}
\end{table}

The fast octree optimization
 indirectly allows us to accelerate NeRF training, as seen in
Fig.~\ref{fig:early_stop},
since we can stop the NeRF-SH training at an earlier time 
for constructing the PlenOctree,
with only a slight degradation in quality.

\section{Results}

\begin{figure*}[]
    \centering
    \includegraphics[width=\linewidth]{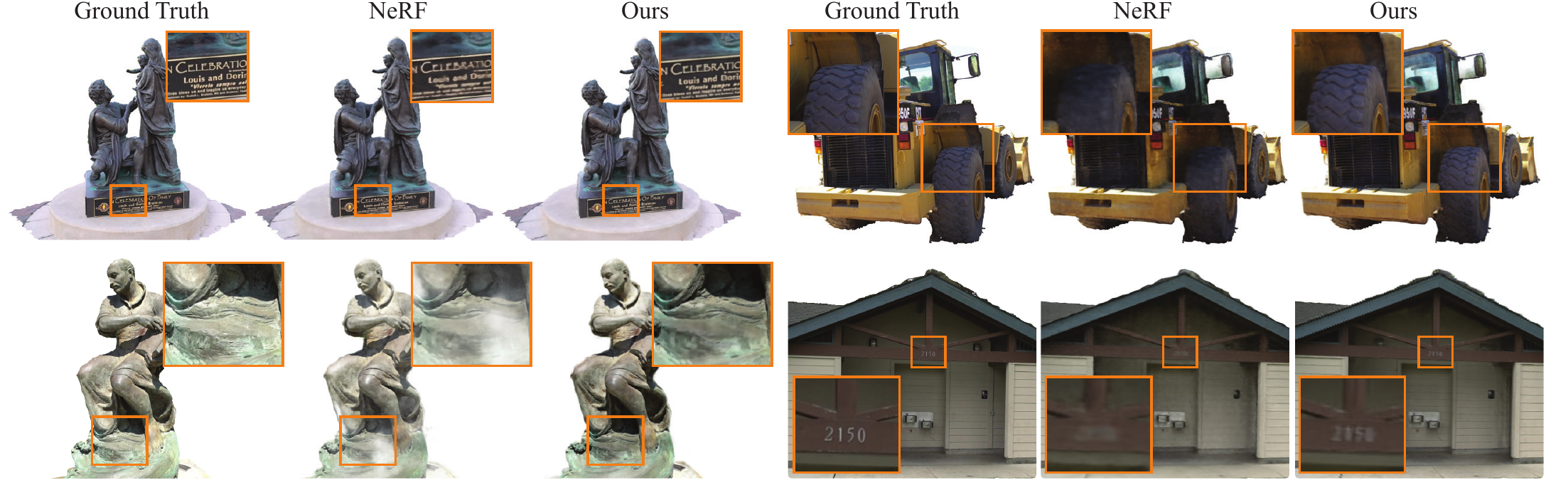}
    \caption{\textbf{Qualitative comparisons on Tanks and Temples.} We compare NeRF and our proposed method. On this dataset, we find that our method better recovers the fine details in the scene. The results are otherwise similar. Additionally, the render time of our method is over 3000$\times$ faster.}
    \label{fig:qual_comparison_tt}
\end{figure*}

\subsection{Experimental Setup}
\inlinepara{Datasets.}
For our experiments, we use the \textit{NeRF-synthetic}~\cite{mildenhall2020} dataset and a subset of the \textit{Tanks and Temples} dataset~\cite{knapitsch2017tanks}. The NeRF-synthetic dataset consists of 8 scenes where each scene has a central object with 100 inward facing cameras distributed randomly on the upper hemisphere. The images are $800\times800$ with provided ground truth camera poses. The Tanks and Temples subset is from NSVF~\cite{nsvf} and contains 5 scenes of real objects captured by an inward facing camera circling the scene.
We use foreground masks provided by the NSVF authors. Each scene contains between 152-384 images of size $1920\times1080$. 

\vspace{0.2em}
\inlinepara{Baselines.}
The principal baseline for our experiments is NeRF~\cite{mildenhall2020};
we report results for both the
original NeRF implementation, denoted \textit{NeRF (original)}
as well as a reimplementation in Jax~\cite{jaxnerf2020github}, denoted simply
\textit{NeRF}, which our \OurNeRF code is based off of.
Unless otherwise stated, all \textit{NeRF} results and timings are from the latter implementation.
We compare also to two recent
papers introducing NeRF accelerations,
neural sparse voxel fields (NSVF)~\cite{nsvf}
and AutoInt~\cite{lindell2020autoint},
as well as two older methods, scene representation networks (SRN)~\cite{sitzmann2019scene} and
Neural Volumes~\cite{lombardi2019neuralvol}.

\subsection{Quality Evaluation}
We evaluate our approach against prior works on the synthetic and real datasets mentioned above. 
The results are in Tables~\ref{tab:synthetic_results} and Table~\ref{tab:tt_results} respectively.
Note that none of the baselines achieve real-time performance; nevertheless,
our quality results are competitive in all cases and better in terms of some metrics.

In Figures~\ref{fig:qual_comparison} and \ref{fig:qual_comparison_tt}, 
we show qualitative examples that demonstrate that our PlenOctree conversion does not perceptually worsen the rendered images compared to NeRF;
rather, we observe that the PlenOctree optimization process \textit{enhances} fine details such as text.
Additionally, we note that our modifications of NeRF to predict spherical function coefficients (\OurNeRF) does not significantly change the performance.

For the SH, we set $\l_{\max} = 3$ (16 components) and $4$ (25 components)
on the synthetic and Tanks \& Temples datasets respectively.
We use $512^3$ grid size in either case.
Please refer to the appendix for training details. The inference time performance
is measured on a Tesla V100 for all methods.  Across both datasets we find that PlenOctree inference is over 3000 times faster than NeRF and at least 30 times faster than all other compared methods. \textit{PlenOctree} performs either best, or second best for all image quality metrics.

\subsection{Speed Trade-off Analysis}
A number of parameters for PlenOctree conversion and rendering can be tuned to trade-off between speed and image quality. In Figure~\ref{fig:time_tradeoff} and Table~\ref{tab:ablation_orf} we compared image accuracy and inference time for four variants of PlenOctree that sweep this trade-off.

\subsection{Indirect Acceleration of NeRF Training}

Since we can efficiently fine-tune the octree on the original training data,
as briefly discussed in \S\ref{subsection:finetune},
we can choose to stop the NeRF-SH training at an earlier time 
 before converting it to a PlenOctree.
Indeed, we have found that the image quality improvements gained during 
fine-tuning can often be greater than continuing to train the \OurNeRF an equivalent amount of time. Therefore it can be more time efficient to stop the \OurNeRF training before it has converged and transition to PlenOctree conversion and fine-tuning.

\begin{figure}[t]
    \centering
    \includegraphics[width=\linewidth]{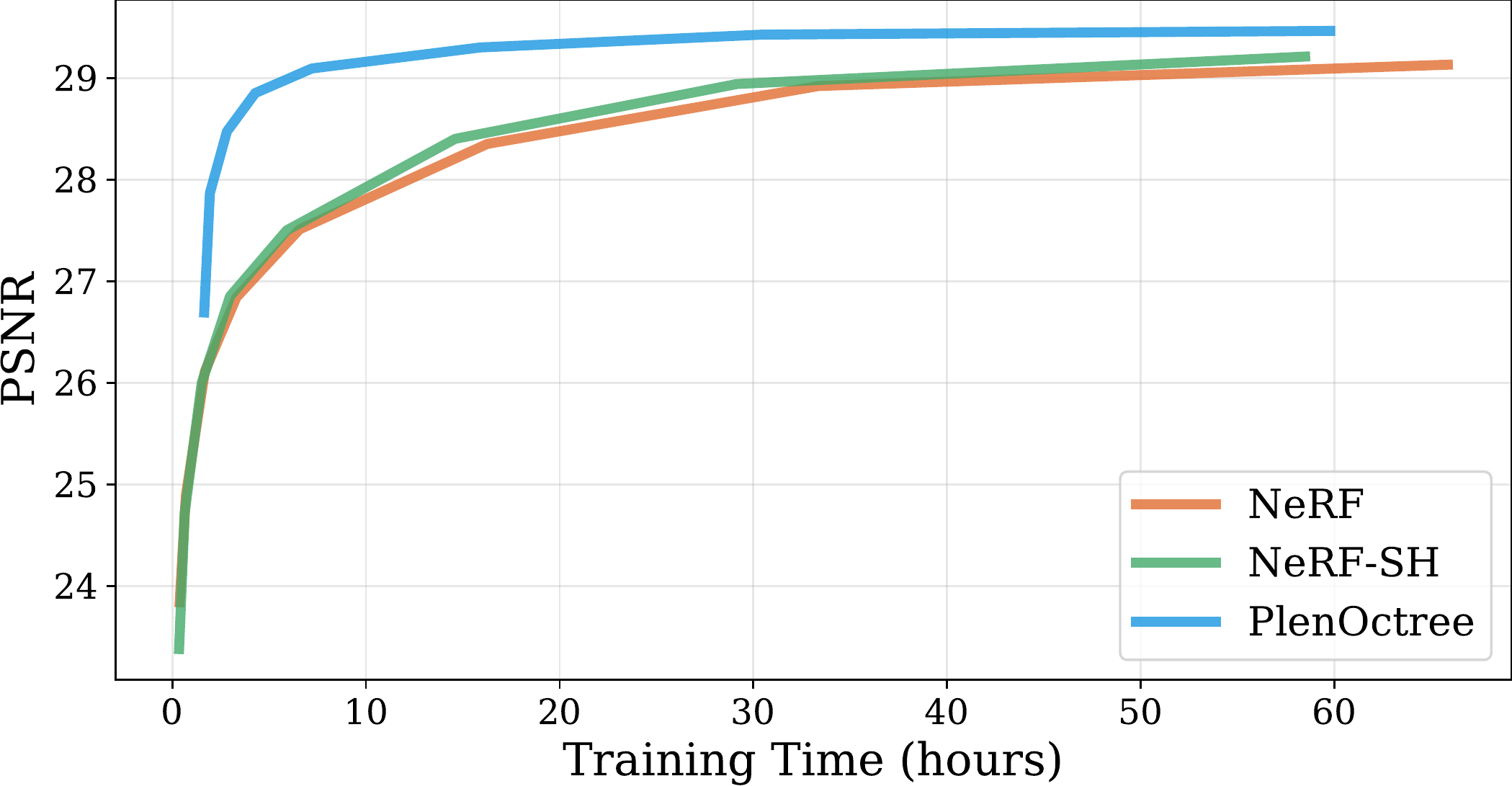}
    \caption{\textbf{Indirect training acceleration.} Training curves for fitting the synthetic NeRF \textit{ship} scene for 2 million iterations. The baseline NeRF model and our \OurNeRF{} model perform similarly during training. We find that by optimizing the PlenOctree converted from \OurNeRF{} checkpoints, we are able to reach a similar quality more quickly. The PlenOctree conversion and fine-tuning adds approximately 1 hour to the training time; despite this, we find that it takes $\sim$16 hours of NeRF training to match the same quality as the PlenOctree model after $\sim$4.5 
    hours.}
    \label{fig:early_stop}
\end{figure}

In Figure~\ref{fig:early_stop}
we compare NeRF and \OurNeRF models trained for 2 million iterations each to
a sequence of PlenOctree models extracted from \OurNeRF checkpoints.
We find that given a time constraint, it is almost always preferable to stop the NeRF training and transition to PlenOctree optimization.

\begin{figure}[t]
    \centering
    \includegraphics[width=\linewidth]{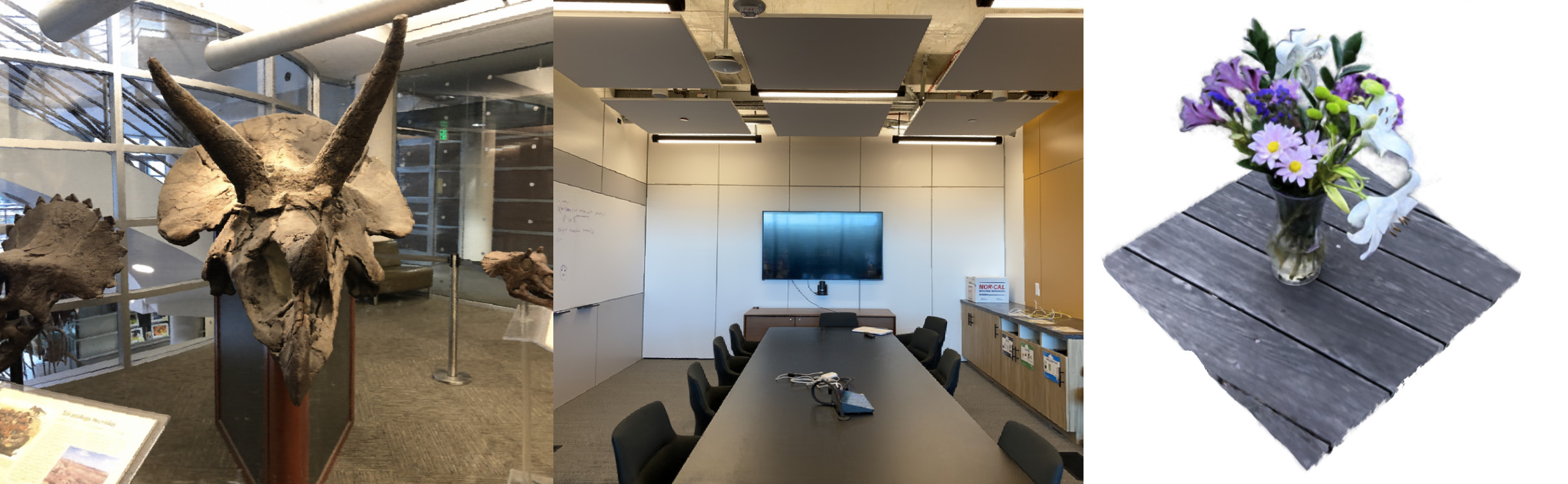}
    \caption{
    \textbf{Qualitative results on further real scenes.}
     We apply \PaperName to the NeRF-360-real and LLFF datasets.
     While our method is not designed for unbounded or forward-facing scenes (where 
     MPIs can be more appropriate),
     it nevertheless performs reasonably well.
     Note for forward-facing scenes, we construct the octree in NDC coordinates.
    }
    \label{fig:realimages}
\end{figure}

\subsection{Real-time and In-browser Applications}
\begin{figure}[t]
    \centering
    \includegraphics[width=\linewidth]{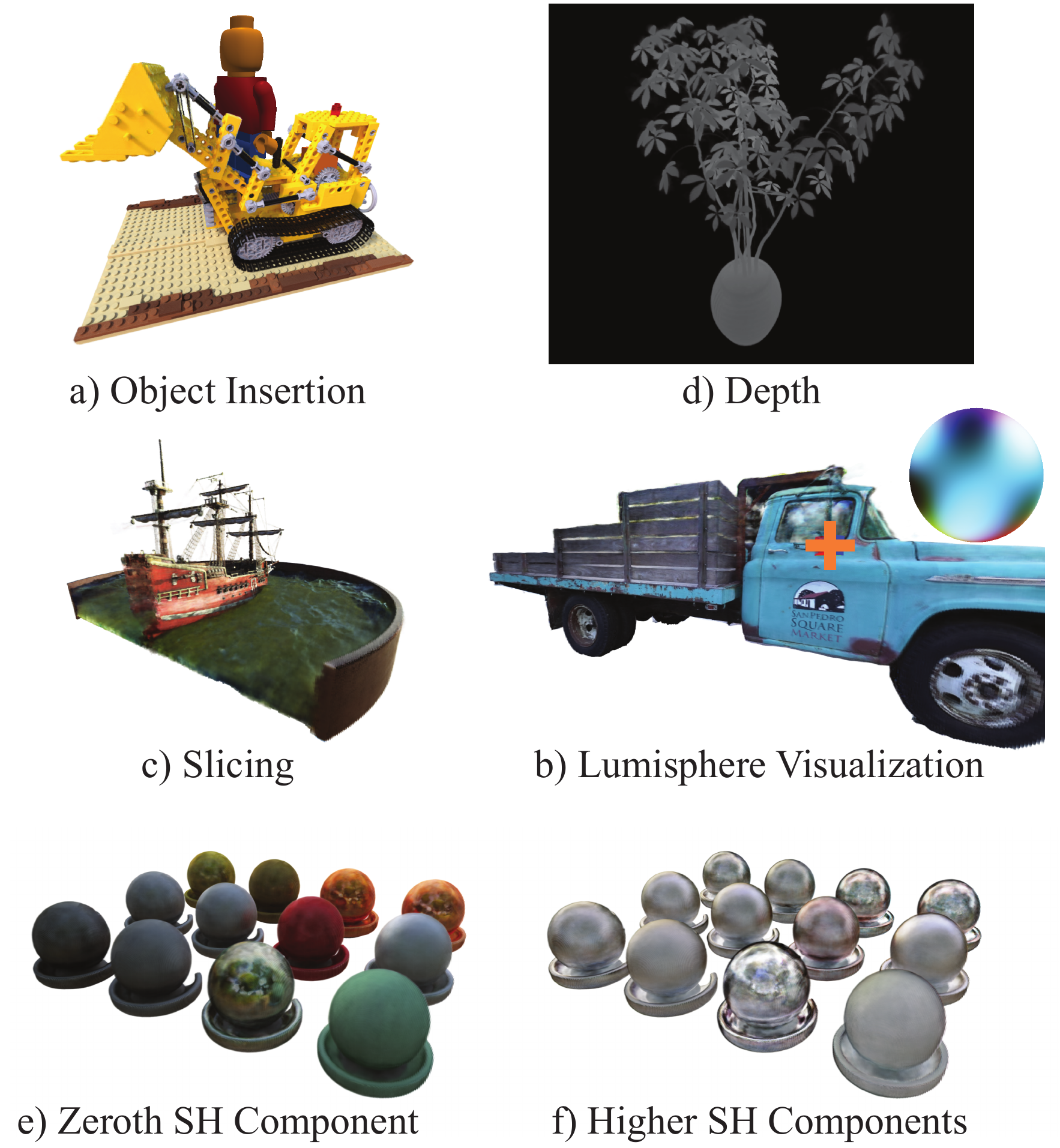}
    \caption{\textbf{Real-time interactive demos.} Set of real-time operations that can be performed on PlenOctree 
    within our interactive viewer. This application 
   will be released to the public.}
    \label{fig:realtime_demo}
\end{figure}

\inlinepara{Interactive demos.}
Within our desktop viewer, we are able to perform a variety of real-time scene operations on the PlenOctree representation. For example, it is possible to insert meshes while maintaining proper occlusion, slice the PlenOctree to visualize a cross-section, or render the depth map to verify the geometry. Other features include probing the radiance distribution at any point in space,
and inspecting subsets of SH components. These examples are demonstrated in Figure~\ref{fig:realtime_demo}. The ability to perform these actions in real-time is beneficial both for interactive entertainment and debugging NeRF-related applications. 

\vspace{0.2em}
\inlinepara{Web renderer.}
We have implemented a web-based renderer
enabling interactive viewing of PlenOctrees in the browser.
This is achieved by rewriting our CUDA-based PlenOctree renderer as a WebGL-compatible fragment shader
and applying compressions to reduce file sizes. Please see the appendix for more information.

\section{Discussion}

We have introduced a new data representation for NeRFs using PlenOctrees, which enables real-time rendering capabilities for arbitrary objects and scenes. Not only can we accelerate the rendering performance of the original NeRF method by more than 3000 times, but we can produce images that are either equal or better quality than NeRF thanks to our hierarchical data structure.
As training time poses another hurdle for adopting NeRFs in practice (taking 1-2 days to fully converge), we also showed that our PlenOctrees can accelerate effective training time for our \OurNeRF{}. Finally, we have implemented an in-browser viewer based on WebGL to demonstrate real-time and 6-DOF rendering capabilities of NeRFs on consumer laptops. 
In the future, our approach may enable virtual online stores in VR, where any products with arbitrary complexity and materials can be visualized in real-time while enabling 6-DOF viewing. 

\paragraph{Limitations and Future Work.}
While we achieve state-of-the-art rendering performance and frame rates, 
the octree representation is much larger than the compact representation of the original NeRF model and has a larger memory footprint.
The average uncompressed octree size for the full model is 1.93 GB on the synthetic dataset and 3.53 GB on the Tanks and Temples dataset.
For online delivery, we use lower-resolution compressed models
which are about 30-120 MB; please
see the appendix for details.
Although already possible in some form (Fig.~\ref{fig:realimages}), applying our method to unbounded and forward-facing scenes optimally requires further work as the data distribution is different for unbounded scenes.
The forward-facing scenes inherently do not support 6-DOF viewing, and we suggest MPIs may be more appropriate in this case~\cite{wizadwongsa2021nex}.

In the future, we plan to explore extensions of our method to enable real-time 6-DOF immersive viewing of large-scale scenes, as well as of dynamic scenes. We believe that real-time rendering of NeRFs has the potential to become a new standard for next-generation AR/VR technologies, as photorealistic 3D content can be digitized as easily as recording 2D videos.

\section*{Acknowledgements}

We thank Vickie Ye and Ben Recht for help discussions as well as reviewing the manuscript, Zejian Wang of Pinscreen for helping with video capture, and BAIR commons for an allocation of GCP credits.

{\small
\bibliographystyle{ieee_fullname}
\bibliography{egbib}
}

\appendix
\section*{{\Large Appendix}}

\section{Additional Results}

\begin{figure*}[]
    \centering
    \vspace{3mm}
    \includegraphics[width=\linewidth]{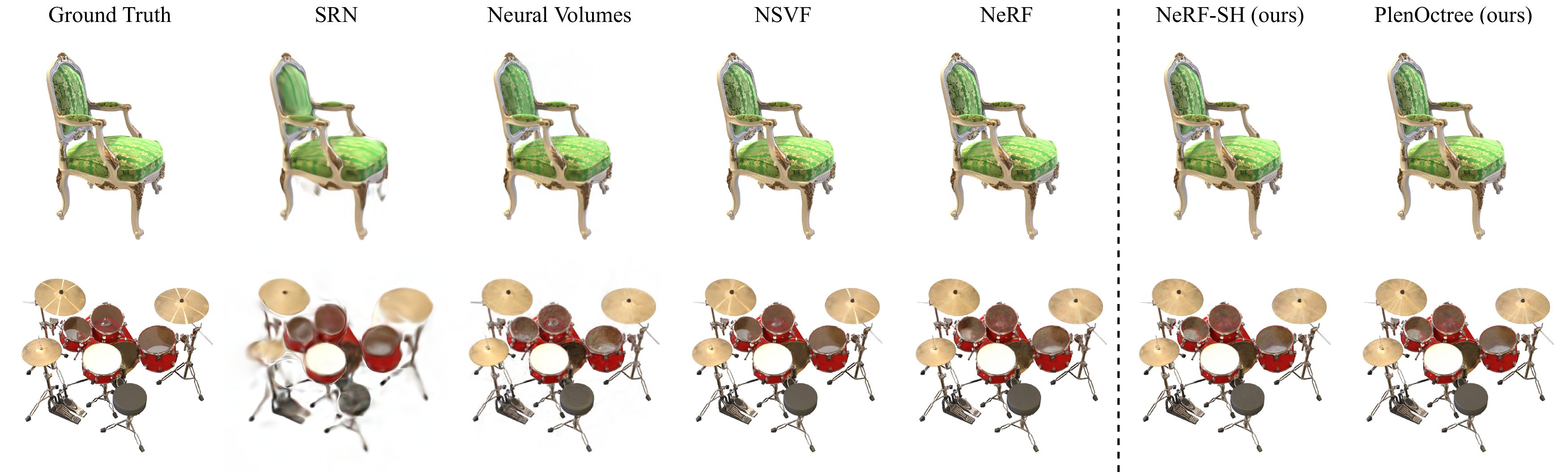}
    \caption{\textbf{Qualitative comparisons on NeRF-synthetic.}}
    \vspace{1mm}
    \label{fig:qual_comparison_syn}
    \vspace{2mm}
\end{figure*}

\subsection{Detailed comparisons}
\begin{figure*}[]
    \centering
    \includegraphics[width=\linewidth]{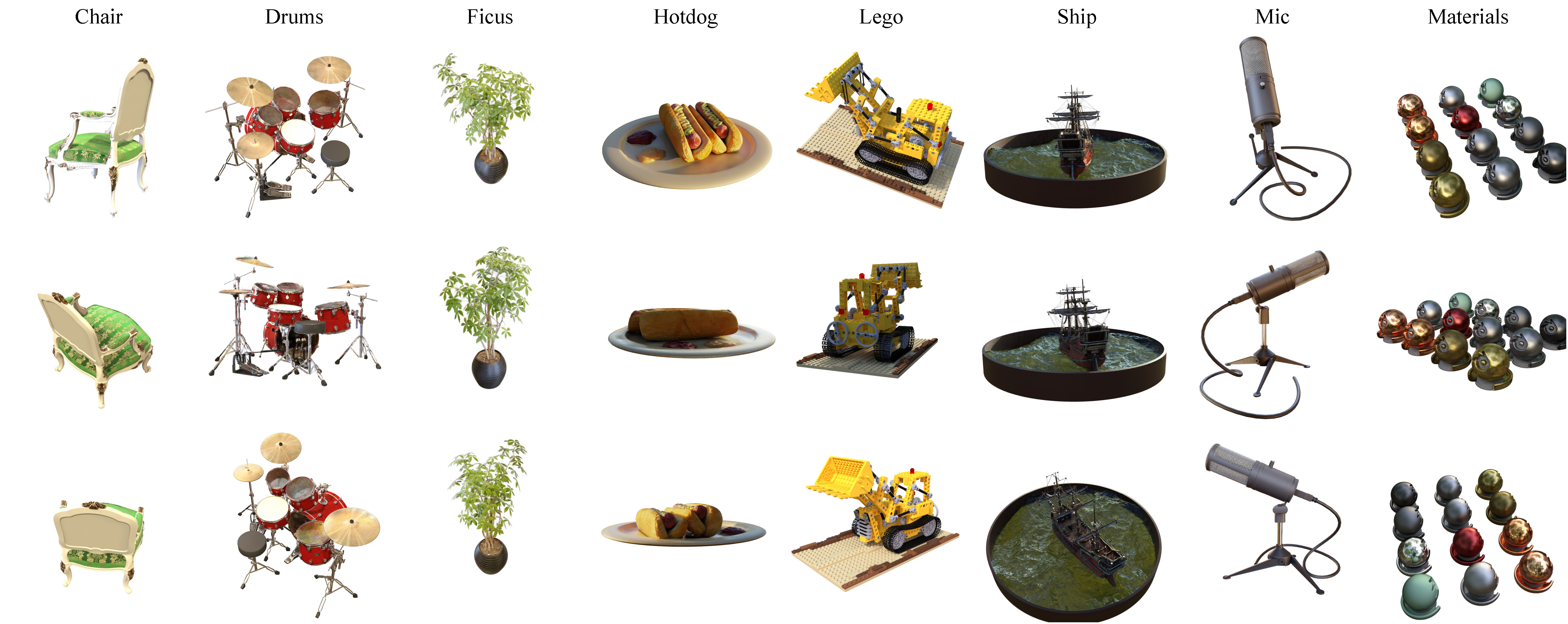}
    \caption{\textbf{More qualitative results of our PlenOctrees on NeRF-synthetic.}}
    \label{fig:results_syn}
\end{figure*}

\begin{figure*}[]
    \centering
    \includegraphics[width=\linewidth]{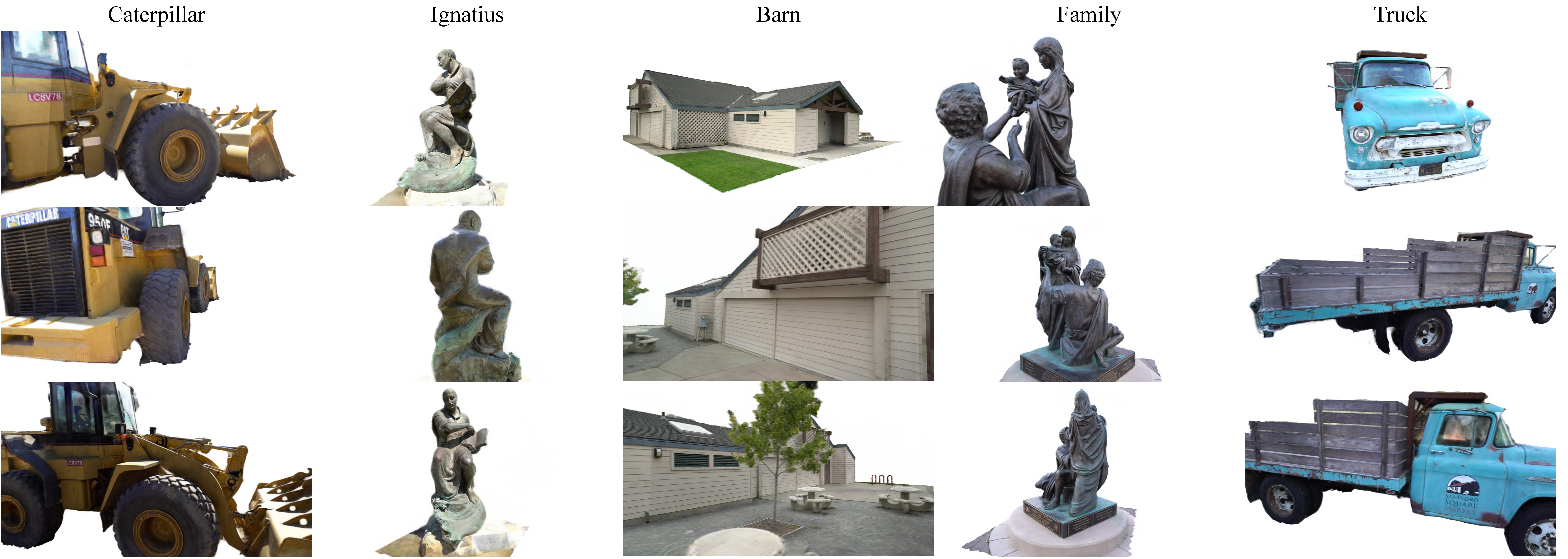} 
    \caption{\textbf{More qualitative results of our PlenOctrees on Tanks\&Temples.}} 
    \label{fig:results_tt}
\end{figure*}

Here we provide further qualitative comparisons with baselines: SRN~\cite{sitzmann2019scene}, Neural Volumes~\cite{lombardi2019neuralvol}, NSVF~\cite{nsvf} in Figure~\ref{fig:qual_comparison_syn}. We show more qualitative results of our method in Figure~\ref{fig:results_syn} and Figure~\ref{fig:results_tt}.  We also report a per-scene breakdown of the quantitative metrics against all approaches in Table~\ref{tab:per_scene_syn},~\ref{tab:per_scene_tt},~\ref{tab:per_scene_tradeoff},~\ref{tab:per_scene_sh_sg}. 

\subsection{Spherical Basis Function Ablation}
We also provide ablation studies on the choice of spherical basis functions. We first ablate the effect on the number of spherical harmonics basis, then we explore the use of a learnable spherical basis functions. All experiments are conducted on NeRF-synthetic dataset and we report the average metric directly after training NeRF with spherical basis functions and after converting it to PlenOctrees  with fine-tuning. 

\paragraph{Number of SH basis functions}
First, we ablate the number of basis functions used for spherical harmonics. 
Average metrics across the NeRF-synthetic dataset are reported
both for the modified NeRF model and the corresponding PlenOctree. We found that switching between $\l_{\max} = 3$ (\textit{SH-16})
and $4$ (\textit{SH-25}) makes very little difference in terms of metrics or visual quality.

\paragraph{Spherical Gaussians}
Furthermore, we also experimented with spherical Gaussians (SGs)~\cite{fisher1953dispersion}, which is another form of spherical basis functions similar to spherical harmonics, but with learnable Gaussian kernels. Please see \S\ref{subsection:sh_sg} for a brief introduction of SHs and SGs. \textit{SG-25} denotes our model using 25 SG components instead of SH, all with learnable lobe axis $\mathbf{p}$ and bandwidth $\lambda$.
However, while this model has marginally better PSNR, the advantage disappears following PlenOctree conversion and fine-tuning. 

\begin{table}[]
    \centering
    \resizebox{\linewidth}{!}{
     \begin{tabular}{@{}l c ccc c ccccc}
\toprule
 & & \multicolumn{3}{c}{NeRF-SH/SG} & & \multicolumn{5}{c}{Converted PlenOctree} \\
\cmidrule{3-5} \cmidrule{7-11}
Basis &  & PSNR $\uparrow$ & SSIM $\uparrow$ & LPIPS $\downarrow$ &  & PSNR $\uparrow$ & SSIM $\uparrow$ & LPIPS $\downarrow$ & GB $\downarrow$ & FPS$\uparrow$ \\
\cmidrule{1-1} \cmidrule{3-5} \cmidrule{7-11}
SH-9 &  & 31.44 & 0.951 & 0.065 &  & 31.45 & 0.956 & 0.056 & 1.00 & 262 \\
SH-16 &  & 31.57 & 0.952 & 0.063 &  & 31.71 & 0.958 & 0.053 & 1.93 & 168 \\
SH-25 &  & 31.56 & 0.951 & 0.063 &  & 31.69 & 0.958 & 0.052 & 2.68 & 128 \\
SG-25 &  & 31.74 & 0.953 & 0.062 &  & 31.63 & 0.958 & 0.052 & 2.26 & 151 \\
\bottomrule
\end{tabular}
    }
    \caption{\textbf{Spherical Basis Function Ablation.} We experiment with various versions of spherical basis functions, including \textit{SH-16}, \textit{SH-25} and \textit{SG-25}.}
    \label{tab:basis_ablation}
\end{table}

\section{Technical Details}
\subsection{Spherical Basis Functions: SH and SG}
\label{subsection:sh_sg}

In the main paper, we used the SH functions without defining their exact form.
Here, 
we provide a brief technical discussion of both spherical harmonics (SH) and spherical Gaussians (SG) for completeness.

\paragraph{Spherical Harmonics.}
The Spherical Harmonics (SH) form a complete basis of functions $\sphere \rightarrow \Complex$.
For $\l \in \mathbb{N} \cup \{0\}$ and
$m \in \{-\l, \ldots, \l\}$, the
SH function of degree $\l$ and order $m$ is defined as:
\begin{equation}
Y_{\l}^{m}(\theta, \phi) = \sqrt{\frac{2\l+1}{4\pi} \frac{(\l-m)!}{(\l+m)!}} P_{\l}^{m}(\cos\theta)e^{im\phi}
\end{equation}
where $P_{\l}^{m}(cos\theta)e^{im\phi}$ are the associated Legendre polynomials.
A real basis of SH $Y_{\l}^m: \sphere \mapsto \Real$
can be defined in terms of its complex analogue $Y_{\l}^{m}: \sphere \mapsto \Complex$ by setting
\begin{equation}
Y_{\l}^m(\theta, \phi) =
  \begin{cases}
    \sqrt{2}(-1)^{m}\operatorname{Im}[Y_{\l}^{|m|}] & \text{if $m < 0$} \\
    Y_{\l}^{0} & \text{if $m = 0$} \\
    \sqrt{2}(-1)^{m}\operatorname{Re}[Y_{\l}^{m}] & \text{if $m > 0$}
  \end{cases}
\end{equation}

Any real spherical function $L: \sphere \rightarrow \Real$ may then be expressed in the SH basis:
\begin{equation}
    L(\direction) = L(\theta, \phi) = \sum_{\l=0}^{\infty} \sum_{m=-\l}^{\l} 
            k_{\l}^m Y_{\l}^m(\theta, \phi)
\end{equation}

\paragraph{Spherical Gaussians.}
Spherical Gaussians (SGs), also known as the von Mises-Fisher distribution~\cite{fisher1953dispersion}, is another form of spherical basis functions that have been widely adopted to approximate spherical functions. Unlike 
SHs, SGs are a learnable basis. A normalized SG is defined as:
\begin{equation}
    G(\direction; \sglobe, \lambda) = e^{\lambda(\direction\cdot\sglobe - 1)}
\end{equation}
Where $\sglobe \in \Real^2$  is the lobe axis, and $\lambda \in \Real$ is the bandwidth (sharpness) of the Gaussian kernel. Due to the varying bandwidths supported by SGs, they are suitable for representing all-frequency
signals such as lighting~\cite{tsai2006all, sloan2002precomputed, li2020inverse}. A spherical function represented using $n$ SGs is formulated as:
\begin{equation}
    L(\direction) \approx \sum_{\l=0}^{n} k_{\l} G_{\l}(\direction; \sglobe, \lambda)
\end{equation}
Where $k_{\l} \in \Real^3$ is the RGB coefficients for each SG.

\subsection{PlenOctree Compression}
The uncompressed PlenOctree file would be unpleasantly time-consuming for users to download for in-browser rendering. Thus, to minimize the size of PlenOctrees for viewing in the browser, we use SH-9 instead of SH-16 or SH-25 and
        apply a looser bounding box, which reduces the number of occupied voxels.
        On top of this, we compress the PlenOctrees directly in the following ways:
\begin{enumerate}
    \item 
    We quantize the SH coefficients in the tree using the popular median-cut algorithm~\cite{mediancut}.
    More specifically,
    the $\sigma$ values are kept as is;
    for each SH basis function, we quantize the RGB coefficients $k_{\l}^m \in \Real^3$
    into $2^{16}$ colors.
    Afterwards, separately for each SH basis function, we store a 
        $2^{16} \times 3$ codebook (as \texttt{float16})
    along with pointers from each tree leaf to a position in the codebook (as \texttt{int16}).
    \item
We compress the entire tree, including pointers, using the standard DEFLATE algorithm from ZLIB~\cite{zlib}.
\end{enumerate}

This process reduces the file size by as much as $20$-$30$ times.
The tree is fully decompressed before it is displayed in the web renderer. We will also release this code.

\subsection{Analytic Derivatives of PlenOctree Rendering}

In this section, we derive the analytic derivatives of the NeRF piecewise constant volume rendering model
for optimizing PlenOctrees directly.
Throughout this section we will consider a fixed ray with a given origin and direction.

\subsubsection{Definitions}
For preciseness, we provide definitions of quantities used in NeRF volume rendering.
The NeRF rendering model 
considers
a ray divided into
$N$ consecutive segments 
with endpoints $\{t_i\}_{i =0}^{N}$, where $t_0$ and $t_N$ are the near and far bounds.
The segments have constant densities $\bm \sigma = (\sigma_0, \ldots, \sigma_{N-1})$ where each $\sigma_i \ge 0$.
If we shine a light of intensity $1$ at $t_i$, then
at the camera position $t_0$, the light intensity is given by
\begin{equation}
T_i(\bm \sigma) := \prod_{j=0}^{i-1} \exp(-\delta_j \sigma_j),
\end{equation}
Where  \(\delta_i := t_{i+1} - t_i\) are segment lengths as in $\S$\ref{section:prelim} of the main paper. 
Note that $T_i$ is also known as the accumulated transmittance from $t_0$ to $t_i$,
and is the same as the definition in \eqref{eqn:quad}. 
It can be shown that this precisely models the absorption 
within each segment
in the piecewise-constant setting.

Let $\tf c = (c_0, \ldots c_{N-1})$
be the color associated with segments $0, \ldots, N-1$, and
$c_N$ be the background light intensity;
each $c_0, \ldots, c_N \in [0,1]^3$ is an RGB color.
We are interested in the derivative of 
the rendered color $\estimate{C}(\bm \sigma, \tf c)$
with respect to $\bm \sigma$ and $\tf c$.
Note $c_N$ (background) is typically considered a hyperparameter.

\subsubsection{Derivation of the Derivatives}
From the original NeRF rendering equation \eqref{eqn:quad},
we can express
the rendered ray color $\estimate{C}(\bm \sigma, \tf c)$ as:
\begin{align}
    \estimate{C}\left(\bm \sigma, \tf c\right) &=
    T_N(\bm \sigma)\,c_N +
    \sum_{i=0}^{N-1} T_i(\bm \sigma) \big(1 - e^{ -\sigma_{i} \delta_{i}}\big) \, c_{i}\\
    &= \sum_{i=0}^{N}
        w_i(\bm \sigma)
        \,
        c_i
\end{align}
Where  \(
w_i(\bm \sigma)
            = T_i(\bm \sigma) \left(1 - \exp(-\sigma_i \delta_i)\right)
            = T_i(\bm \sigma) -
        T_{i+1}(\bm \sigma)
       \) 
        are segment weights,
        and $w_N(\bm \sigma) = T_N(\bm \sigma)$.\footnote{
        Note that the background
        color $c_N$ was omitted in equation \eqref{eqn:quad} of the main paper for simplicity.
}

\paragraph{Color derivative.} Since the rendered color are a convex combination of the segment colors, it's immediately clear that
\begin{equation}
\frac{\partial \estimate{C}}{\partial c_i}\left(\bm \sigma, \tf c\right)
    = w_i(\bm \sigma)
\end{equation}
Handling spherical harmonics colors is straightforward by applying the chain rule, noting that the SH 
basis function values are constant across the ray.

\paragraph{Density derivative.} This is slightly more tricky. We can write the derivative wrt.~$\sigma_i$,
\begin{equation}
\frac{\partial \estimate{C}}{\partial \sigma_i}\left(\bm \sigma, \tf c\right)
    =
        c_N\;
        \frac{\partial T_N}{\partial \sigma_i}
        +
    \sum_{k=0}^{N-1}c_k\;
    \left(
    \frac{\partial T_k}{\partial \sigma_i} 
    -
    \frac{\partial T_{k+1}}{\partial \sigma_i} 
    \right)
    \label{eqn:dendiv}
\end{equation}
Where the derivative of the intensity $T_k$, is
\begin{align}
\frac{\partial T_k}{\partial \sigma_i}
(\bm \sigma)
    &=
    \frac{\partial}{\partial \sigma_i}
    \left[
        \prod_{j=0}^{k-1}
            e^{-\delta_j \sigma_j}
    \right]\\
    &=
    -\delta_i
    \left[
        \prod_{j=0}^{k-1}
            e^{-\delta_j \sigma_j}
            \right]
            \,
    1_{k> i}\\
    &=
    - \delta_i T_k (\bm \sigma) \,
    1_{k> i}
\end{align}

$1_{k>i}$ denotes an indicator function whose value is $1$ if $k > i$ or $0$ else.
Basically, we can delete any $T_k$ for $k \le i$ from the original expression,
then multiply by $-\delta_i$.
Therefore we can simplify
    \eqref{eqn:dendiv} as follows
\begin{equation}
\frac{\partial \estimate{C}}{\partial \sigma_i}
(\bm \sigma, \tf c)
    =
    \delta_i \left[
        c_i T_{i+1}(\bm \sigma)
        -
    \sum_{k=i+1}^{N}c_k\;
    w_k(\bm \sigma)
    \right]
\end{equation}

\paragraph{Remark.}
Within the PlenOctree renderer, 
this gradient can be computed in two rendering passes; the second pass is needed
due to dependency on ``future'' weights and colors not seen by the ray marching process. The first pass store
\(
    \sum_{k=0}^{N}c_k\;
    w_k(\bm \sigma)
\), then subtracting a prefix from it. The overhead is still relatively small,
and auxiliary memory use is constant.

If there are multiple colors, we simply add the density derivatives over all of them.
In practice, usually the network outputs
$\tilde \sigma_i \in\Real$ and we set
$\sigma_i = (\tilde \sigma_i)_+$,
so we also need to take care of setting the gradient to $0$ if $\tilde \sigma_i\le 0$.

\subsection{NeRF-SH Training Details}

Our \OurNeRF model is built upon a Jax reimplementation of NeRF~\cite{jaxnerf2020github}. In our experiments, we use a batch size of 1024 rays, each with 64 sampled points in the coarse volume and 128 additional sampled points in the fine volume.  The model is optimized with the Adam optimizer~\cite{kingma2014adam} using a learning rate that starts at $5 \times 10^{-4}$ and decays exponentially to $5 \times 10^{-6}$ over the training process. All of our models are trained for 2M iterations under the same protocol. Training takes around 50 hours to converge for each model on a single NVIDIA V100 GPU.

\subsection{PlenOctree Optimization Details}

After converting the \OurNeRF model into a PlenOctree, we further optimize the PlenOctree on the training set with SGD using the NeRF loss; note we no longer apply the sparsity loss here since the octree is already sparse. For NeRF-synthetic dataset, we use a constant $1 \times 10^{7}$ learning rate and optimize for maximum 80 epochs. For Tanks\&Temples dataset, we set the learning rate to $1.5 \times 10^{6}$ and the maximum epochs to 40. We applied early stopping for the optimization process by monitoring the PSNR on the validation set\footnote{For Tanks\&Temples dataset, we hold out 10\% of the training set as validation set only for PlenOctree optimization.}. On average it takes around 10 minutes to finish the PlenOctree optimization for each scene on a single NVIDIA V100 GPU. The entire optimization process is done in \texttt{float32} for stability, but after it we storage the PlenOctree with \texttt{float16} to reduce the model size.

\begin{table*}[t]
    \centering
    \resizebox{0.95\linewidth}{!}{
\begin{tabular}{@{}l c ccccccccc}
\multicolumn{11}{c}{PSNR $\uparrow$}
\vspace{0.1em}\\
\toprule
 &  & Chair & Drums & Ficus & Hotdog & Lego & Materials & Mic & Ship & Mean \\
\cmidrule{1-1} \cmidrule{3-11}
NeRF (original) &  & 33.00 & 25.01 & 30.13 & 36.18 & 32.54 & 29.62 & 32.91 & 28.65 & 31.01 \\
NeRF &  & 34.08 & 25.03 & 30.43 & 36.92 & 33.28 & 29.91 & 34.53 & 29.36 & 31.69 \\
SRN &  & 26.96 & 17.18 & 20.73 & 26.81 & 20.85 & 18.09 & 26.85 & 20.60 & 22.26 \\
Neural Volumes &  & 28.33 & 22.58 & 24.79 & 30.71 & 26.08 & 24.22 & 27.78 & 23.93 & 26.05 \\
NSVF &  & 33.19 & 25.18 & 31.23 & 37.14 & 32.29 & 32.68 & 34.27 & 27.93 & 31.75 \\
\cmidrule{1-1} \cmidrule{3-11}
NeRF-SH &  & 33.98 & 25.17 & 30.72 & 36.75 & 32.77 & 29.95 & 34.04 & 29.21 & 31.57 \\
PlenOctree from NeRF-SH &  & 33.19 & 25.01 & 30.56 & 36.15 & 32.12 & 29.56 & 33.01 & 28.58 & 31.02 \\
PlenOctree after fine-tuning &  & 34.66 & 25.31 & 30.79 & 36.79 & 32.95 & 29.76 & 33.97 & 29.42 & 31.71 \\
\bottomrule
\end{tabular}
}
\bigskip

\resizebox{0.95\linewidth}{!}{
\begin{tabular}{@{}l c ccccccccc}
\multicolumn{11}{c}{SSIM $\uparrow$}
\vspace{0.1em}\\
\toprule
 &  & Chair & Drums & Ficus & Hotdog & Lego & Materials & Mic & Ship & Mean \\
\cmidrule{1-1} \cmidrule{3-11}
NeRF (original) &  & 0.967 & 0.925 & 0.964 & 0.974 & 0.961 & 0.949 & 0.980 & 0.856 & 0.947 \\
NeRF &  & 0.975 & 0.925 & 0.967 & 0.979 & 0.968 & 0.952 & 0.987 & 0.868 & 0.953 \\
SRN &  & 0.910 & 0.766 & 0.849 & 0.923 & 0.809 & 0.808 & 0.947 & 0.757 & 0.846 \\
Neural Volumes &  & 0.916 & 0.873 & 0.910 & 0.944 & 0.880 & 0.888 & 0.946 & 0.784 & 0.893 \\
NSVF &  & 0.968 & 0.931 & 0.960 & 0.987 & 0.973 & 0.854 & 0.980 & 0.973 & 0.953 \\
\cmidrule{1-1} \cmidrule{3-11}
NeRF-SH &  & 0.974 & 0.927 & 0.968 & 0.978 & 0.966 & 0.951 & 0.985 & 0.866 & 0.952 \\
PlenOctree from NeRF-SH &  & 0.970 & 0.927 & 0.968 & 0.977 & 0.965 & 0.953 & 0.983 & 0.863 & 0.951 \\
PlenOctree after fine-tuning &  & 0.981 & 0.933 & 0.970 & 0.982 & 0.971 & 0.955 & 0.987 & 0.884 & 0.958 \\
\bottomrule
\end{tabular}
}
\bigskip

\resizebox{0.95\linewidth}{!}{
\begin{tabular}{@{}l c ccccccccc}
\multicolumn{11}{c}{LPIPS $\downarrow$}
\vspace{0.1em}\\
\toprule
 &  & Chair & Drums & Ficus & Hotdog & Lego & Materials & Mic & Ship & Mean \\
\cmidrule{1-1} \cmidrule{3-11}
NeRF (original) &  & 0.046 & 0.091 & 0.044 & 0.121 & 0.050 & 0.063 & 0.028 & 0.206 & 0.081 \\
NeRF &  & 0.035 & 0.085 & 0.038 & 0.079 & 0.040 & 0.060 & 0.019 & 0.185 & 0.068 \\
SRN &  & 0.106 & 0.267 & 0.149 & 0.100 & 0.200 & 0.174 & 0.063 & 0.299 & 0.170 \\
Neural Volumes &  & 0.109 & 0.214 & 0.162 & 0.109 & 0.175 & 0.130 & 0.107 & 0.276 & 0.160 \\
NSVF &  & 0.043 & 0.069 & 0.017 & 0.025 & 0.029 & 0.021 & 0.010 & 0.162 & 0.047 \\
\cmidrule{1-1} \cmidrule{3-11}
NeRF-SH &  & 0.037 & 0.087 & 0.039 & 0.041 & 0.041 & 0.060 & 0.021 & 0.177 & 0.063 \\
PlenOctree from NeRF-SH &  & 0.039 & 0.088 & 0.038 & 0.044 & 0.046 & 0.063 & 0.023 & 0.189 & 0.066 \\
PlenOctree after fine-tuning &  & 0.022 & 0.076 & 0.038 & 0.032 & 0.034 & 0.059 & 0.017 & 0.144 & 0.053 \\
\bottomrule
\end{tabular}
}
\bigskip

\resizebox{0.95\linewidth}{!}{
\begin{tabular}{@{}l c ccccccccc}
\multicolumn{11}{c}{FPS $\uparrow$}
\vspace{0.1em}\\
\toprule
 &  & Chair & Drums & Ficus & Hotdog & Lego & Materials & Mic & Ship & Mean \\
\cmidrule{1-1} \cmidrule{3-11}
NeRF (original) &  & 0.023 & 0.023 & 0.023 & 0.023 & 0.023 & 0.023 & 0.023 & 0.023 & 0.023 \\
NeRF &  & 0.045 & 0.045 & 0.045 & 0.045 & 0.045 & 0.045 & 0.045 & 0.045 & 0.045 \\
SRN &  & 0.909 & 0.909 & 0.909 & 0.909 & 0.909 & 0.909 & 0.909 & 0.909 & 0.909 \\
Neural Volumes &  & 3.330 & 3.330 & 3.330 & 3.330 & 3.330 & 3.330 & 3.330 & 3.330 & 3.330 \\
NSVF &  & 1.044 & 0.735 & 0.597 & 0.660 & 0.633 & 0.517 & 1.972 & 0.362 & 0.815 \\
\cmidrule{1-1} \cmidrule{3-11}
NeRF-SH &  & 0.051 & 0.051 & 0.051 & 0.051 & 0.051 & 0.051 & 0.051 & 0.051 & 0.051 \\
PlenOctree &  & 352.4 & 175.9 & 85.6 & 95.5 & 186.8 & 64.2 & 324.9 & 56.0 & 167.7 \\
\bottomrule
\end{tabular}
}
    \vspace{0.1em}
    \caption{\textbf{Per-scene quantitive results on NeRF-synthetic dataset.}
    }
    \label{tab:per_scene_syn}
\end{table*}

\begin{table*}[t]
    \centering
    \resizebox{0.75\linewidth}{!}{
\begin{tabular}{@{}l c cccccc}
\multicolumn{8}{c}{PSNR $\uparrow$}
\vspace{0.1em}\\
\toprule
 &  & Barn & Caterpillar & Family & Ignatius & Truck & Mean \\
\cmidrule{1-1} \cmidrule{3-8}
NeRF (original) &  & 24.05 & 23.75 & 30.29 & 25.43 & 25.36 & 25.78 \\
NeRF &  & 27.39 & 25.24 & 32.47 & 27.95 & 26.66 & 27.94 \\
SRN &  & 22.44 & 21.14 & 27.57 & 26.70 & 22.62 & 24.09 \\
Neural Volumes &  & 20.82 & 20.71 & 28.72 & 26.54 & 21.71 & 23.70 \\
NSVF &  & 27.16 & 26.44 & 33.58 & 27.91 & 26.92 & 28.40 \\
\cmidrule{1-1} \cmidrule{3-8}
NeRF-SH &  & 27.05 & 25.06 & 32.28 & 28.06 & 26.66 & 27.82 \\
PlenOctree from NeRF-SH &  & 25.78 & 24.80 & 32.04 & 27.92 & 26.15 & 27.34 \\
PlenOctree after fine-tuning &  & 26.80 & 25.29 & 32.85 & 28.19 & 26.83 & 27.99 \\
\bottomrule
\end{tabular}
}
\bigskip

\resizebox{0.75\linewidth}{!}{
\begin{tabular}{@{}l c cccccc}
\multicolumn{8}{c}{SSIM $\uparrow$}
\vspace{0.1em}\\
\toprule
 &  & Barn & Caterpillar & Family & Ignatius & Truck & Mean \\
\cmidrule{1-1} \cmidrule{3-8}
NeRF (original) &  & 0.750 & 0.860 & 0.932 & 0.920 & 0.860 & 0.864 \\
NeRF &  & 0.842 & 0.892 & 0.951 & 0.940 & 0.896 & 0.904 \\
SRN &  & 0.741 & 0.834 & 0.908 & 0.920 & 0.832 & 0.847 \\
Neural Volumes &  & 0.721 & 0.819 & 0.916 & 0.922 & 0.793 & 0.834 \\
NSVF &  & 0.823 & 0.900 & 0.954 & 0.930 & 0.895 & 0.900 \\
\cmidrule{1-1} \cmidrule{3-8}
NeRF-SH &  & 0.838 & 0.891 & 0.949 & 0.940 & 0.895 & 0.902 \\
PlenOctree from NeRF-SH &  & 0.820 & 0.889 & 0.948 & 0.940 & 0.889 & 0.897 \\
PlenOctree after fine-tuning &  & 0.856 & 0.907 & 0.962 & 0.948 & 0.914 & 0.917 \\
\bottomrule
\end{tabular}
}
\bigskip

\resizebox{0.75\linewidth}{!}{
\begin{tabular}{@{}l c cccccc}
\multicolumn{8}{c}{LPIPS $\downarrow$}
\vspace{0.1em}\\
\toprule
 &  & Barn & Caterpillar & Family & Ignatius & Truck & Mean \\
\cmidrule{1-1} \cmidrule{3-8}
NeRF (original) &  & 0.395 & 0.196 & 0.098 & 0.111 & 0.192 & 0.198 \\
NeRF &  & 0.286 & 0.189 & 0.092 & 0.102 & 0.173 & 0.168 \\
SRN &  & 0.448 & 0.278 & 0.134 & 0.128 & 0.266 & 0.251 \\
Neural Volumes &  & 0.479 & 0.280 & 0.111 & 0.117 & 0.312 & 0.260 \\
NSVF &  & 0.307 & 0.141 & 0.063 & 0.106 & 0.148 & 0.153 \\
\cmidrule{1-1} \cmidrule{3-8}
NeRF-SH &  & 0.291 & 0.185 & 0.091 & 0.091 & 0.175 & 0.167 \\
PlenOctree from NeRF-SH &  & 0.296 & 0.188 & 0.094 & 0.092 & 0.180 & 0.170 \\
PlenOctree after fine-tuning &  & 0.226 & 0.148 & 0.069 & 0.080 & 0.130 & 0.131 \\
\bottomrule
\end{tabular}
}
\bigskip

\resizebox{0.75\linewidth}{!}{
\begin{tabular}{@{}l c cccccc}
\multicolumn{8}{c}{FPS $\uparrow$}
\vspace{0.1em}\\
\toprule
 &  & Barn & Caterpillar & Family & Ignatius & Truck & Mean \\
\cmidrule{1-1} \cmidrule{3-8}
NeRF (original) &  & 0.007 & 0.007 & 0.007 & 0.007 & 0.007 & 0.007 \\
NeRF &  & 0.013 & 0.013 & 0.013 & 0.013 & 0.013 & 0.013 \\
SRN &  & 0.250 & 0.250 & 0.250 & 0.250 & 0.250 & 0.250 \\
Neural Volumes &  & 1.000 & 1.000 & 1.000 & 1.000 & 1.000 & 1.000 \\
NSVF &  & 10.74 & 5.415 & 2.625 & 6.062 & 5.886 & 6.146 \\
\cmidrule{1-1} \cmidrule{3-8}
NeRF-SH &  & 0.015 & 0.015 & 0.015 & 0.015 & 0.015 & 0.015 \\
PlenOctree (ours) &  & 46.94 & 54.00 & 32.33 & 15.67 & 62.16 & 42.22 \\
\bottomrule
\end{tabular}
}
    \vspace{0.1em}
    \caption{\textbf{Per-scene quantitive results on Tanks\&Temples dataset.}
    }
    \label{tab:per_scene_tt}
\end{table*}

\begin{table*}[t]
    \centering
    \begin{tabular}{@{}l c ccccccccc}
\multicolumn{11}{c}{PSNR $\uparrow$}
\vspace{0.1em}\\
\toprule
 &  & Chair & Drums & Ficus & Hotdog & Lego & Materials & Mic & Ship & Mean \\
\cmidrule{1-1} \cmidrule{3-11}
Ours-1.9G &  & 34.66 & 25.31 & 30.79 & 36.79 & 32.95 & 29.76 & 33.97 & 29.42 & 31.71 \\
Ours-1.4G &  & 34.66 & 25.30 & 30.82 & 36.36 & 32.96 & 29.75 & 33.98 & 29.29 & 31.64 \\
Ours-0.4G &  & 32.92 & 24.82 & 30.07 & 36.06 & 31.61 & 28.89 & 32.19 & 29.04 & 30.70 \\
Ours-0.3G &  & 32.03 & 24.10 & 29.42 & 34.46 & 30.25 & 28.44 & 30.78 & 27.36 & 29.60 \\
\bottomrule
\end{tabular}
\bigskip

\begin{tabular}{@{}l c ccccccccc}
\multicolumn{11}{c}{GB $\downarrow$}
\vspace{0.1em}\\
\toprule
 &  & Chair & Drums & Ficus & Hotdog & Lego & Materials & Mic & Ship & Mean \\
\cmidrule{1-1} \cmidrule{3-11}
Ours-1.9G &  & 0.830 & 1.240 & 1.791 & 2.674 & 2.067 & 3.682 & 0.442 & 2.689 & 1.93 \\
Ours-1.4G &  & 0.671 & 0.852 & 0.943 & 1.495 & 1.421 & 3.060 & 0.569 & 1.881 & 1.36 \\
Ours-0.4G &  & 0.176 & 0.350 & 0.287 & 0.419 & 0.499 & 0.295 & 0.327 & 1.195 & 0.44 \\
Ours-0.3G &  & 0.131 & 0.183 & 0.286 & 0.403 & 0.340 & 0.503 & 0.159 & 0.381 & 0.30 \\
\bottomrule
\end{tabular}
\bigskip

\begin{tabular}{@{}l c ccccccccc}
\multicolumn{11}{c}{FPS $\uparrow$}
\vspace{0.1em}\\
\toprule
 &  & Chair & Drums & Ficus & Hotdog & Lego & Materials & Mic & Ship & Mean \\
\cmidrule{1-1} \cmidrule{3-11}
Ours-1.9G &  & 352.4 & 175.9 & 85.6 & 95.5 & 186.8 & 64.2 & 324.9 & 56.0 & 167.7 \\
Ours-1.4G &  & 399.7 & 222.2 & 147.3 & 163.5 & 247.9 & 68.0 & 393.8 & 75.4 & 214.7 \\
Ours-0.4G &  & 639.6 & 290.0 & 208.7 & 273.5 & 339.0 & 268.0 & 522.6 & 86.7 & 328.5 \\
Ours-0.3G &  & 767.6 & 424.1 & 203.8 & 271.7 & 443.6 & 189.1 & 796.4 & 181.1 & 409.7 \\
\bottomrule
\end{tabular}
    \vspace{0.1em}
    \caption{\textbf{Per-scene quantitive results on PlenOctree conversion ablations.}
    }
    \label{tab:per_scene_tradeoff}
\end{table*}

\begin{table*}[t]
    \centering
    \begin{tabular}{@{}l c ccccccccc}
\multicolumn{11}{c}{PSNR $\uparrow$}
\vspace{0.1em}\\
\toprule
 &  & Chair & Drums & Ficus & Hotdog & Lego & Materials & Mic & Ship & Mean \\
\cmidrule{1-1} \cmidrule{3-11}
NeRF-SH9 &  & 33.88 & 25.24 & 30.69 & 36.68 & 32.73 & 29.53 & 33.68 & 29.11 & 31.44 \\
NeRF-SH16 &  & 33.98 & 25.17 & 30.72 & 36.75 & 32.77 & 29.95 & 34.04 & 29.21 & 31.57 \\
NeRF-SH25 &  & 34.01 & 25.10 & 30.52 & 36.83 & 32.76 & 30.06 & 34.08 & 29.11 & 31.56 \\
NeRF-SG25 &  & 34.08 & 25.40 & 31.21 & 36.92 & 32.93 & 29.77 & 34.31 & 29.28 & 31.74 \\
\cmidrule{1-1} \cmidrule{3-11}
PlenOctree-SH9 &  & 34.38 & 25.34 & 30.72 & 36.68 & 32.79 & 29.16 & 33.23 & 29.28 & 31.45 \\
PlenOctree-SH16 &  & 34.66 & 25.31 & 30.79 & 36.79 & 32.95 & 29.76 & 33.97 & 29.42 & 31.71 \\
PlenOctree-SH25 &  & 34.72 & 25.32 & 30.68 & 36.96 & 32.85 & 29.79 & 33.90 & 29.29 & 31.69 \\
PlenOctree-SG25 &  & 34.37 & 25.52 & 31.16 & 36.67 & 32.98 & 29.41 & 33.63 & 29.32 & 31.63 \\
\bottomrule
\end{tabular}
\bigskip

\begin{tabular}{@{}l c ccccccccc}
\multicolumn{11}{c}{SSIM $\uparrow$}
\vspace{0.1em}\\
\toprule
 &  & Chair & Drums & Ficus & Hotdog & Lego & Materials & Mic & Ship & Mean \\
\cmidrule{1-1} \cmidrule{3-11}
NeRF-SH9 &  & 0.973 & 0.928 & 0.968 & 0.978 & 0.966 & 0.948 & 0.984 & 0.864 & 0.951 \\
NeRF-SH16 &  & 0.974 & 0.927 & 0.968 & 0.978 & 0.966 & 0.951 & 0.985 & 0.866 & 0.952 \\
NeRF-SH25 &  & 0.973 & 0.926 & 0.967 & 0.978 & 0.966 & 0.952 & 0.985 & 0.864 & 0.951 \\
NeRF-SG25 &  & 0.974 & 0.930 & 0.971 & 0.978 & 0.967 & 0.951 & 0.986 & 0.867 & 0.953 \\
\cmidrule{1-1} \cmidrule{3-11}
PlenOctree-SH9 &  & 0.980 & 0.934 & 0.970 & 0.982 & 0.970 & 0.950 & 0.984 & 0.881 & 0.956 \\
PlenOctree-SH16 &  & 0.981 & 0.933 & 0.970 & 0.982 & 0.971 & 0.955 & 0.987 & 0.884 & 0.958 \\
PlenOctree-SH25 &  & 0.981 & 0.935 & 0.971 & 0.983 & 0.971 & 0.955 & 0.987 & 0.883 & 0.958 \\
PlenOctree-SG25 &  & 0.980 & 0.937 & 0.973 & 0.982 & 0.972 & 0.953 & 0.986 & 0.883 & 0.958 \\
\bottomrule
\end{tabular}
\bigskip

\begin{tabular}{@{}l c ccccccccc}
\multicolumn{11}{c}{LPIPS $\downarrow$}
\vspace{0.1em}\\
\toprule
 &  & Chair & Drums & Ficus & Hotdog & Lego & Materials & Mic & Ship & Mean \\
\cmidrule{1-1} \cmidrule{3-11}
NeRF-SH9 &  & 0.037 & 0.086 & 0.043 & 0.044 & 0.042 & 0.063 & 0.023 & 0.180 & 0.065 \\
NeRF-SH16 &  & 0.037 & 0.087 & 0.039 & 0.041 & 0.041 & 0.060 & 0.021 & 0.177 & 0.063 \\
NeRF-SH25 &  & 0.038 & 0.087 & 0.039 & 0.040 & 0.041 & 0.061 & 0.021 & 0.179 & 0.063 \\
NeRF-SG25 &  & 0.036 & 0.083 & 0.034 & 0.042 & 0.041 & 0.060 & 0.020 & 0.176 & 0.062 \\
\cmidrule{1-1} \cmidrule{3-11}
PlenOctree-SH9 &  & 0.023 & 0.075 & 0.041 & 0.034 & 0.036 & 0.068 & 0.025 & 0.146 & 0.056 \\
PlenOctree-SH16 &  & 0.022 & 0.076 & 0.038 & 0.032 & 0.034 & 0.059 & 0.017 & 0.144 & 0.053 \\
PlenOctree-SH25 &  & 0.023 & 0.072 & 0.036 & 0.031 & 0.034 & 0.060 & 0.017 & 0.145 & 0.052 \\
PlenOctree-SG25 &  & 0.023 & 0.069 & 0.034 & 0.033 & 0.033 & 0.064 & 0.019 & 0.144 & 0.052 \\
\bottomrule
\end{tabular}
\bigskip

\begin{tabular}{@{}l c ccccccccc}
\multicolumn{11}{c}{GB $\downarrow$}
\vspace{0.1em}\\
\toprule
 &  & Chair & Drums & Ficus & Hotdog & Lego & Materials & Mic & Ship & Mean \\
\cmidrule{1-1} \cmidrule{3-11}
PlenOctree-SH9 &  & 0.45 & 0.67 & 1.15 & 1.27 & 1.16 & 1.48 & 0.16 & 1.67 & 1.00 \\
PlenOctree-SH16 &  & 0.83 & 1.24 & 1.79 & 2.67 & 2.07 & 3.68 & 0.44 & 2.69 & 1.93 \\
PlenOctree-SH25 &  & 1.30 & 1.97 & 2.57 & 3.80 & 3.61 & 4.04 & 0.55 & 3.61 & 2.68 \\
PlenOctree-SG25 &  & 1.03 & 1.68 & 2.43 & 2.66 & 2.66 & 4.44 & 0.49 & 2.71 & 2.26 \\
\bottomrule
\end{tabular}
\bigskip

\begin{tabular}{@{}l c ccccccccc}
\multicolumn{11}{c}{FPS $\uparrow$}
\vspace{0.1em}\\
\toprule
 &  & Chair & Drums & Ficus & Hotdog & Lego & Materials & Mic & Ship & Mean \\
\cmidrule{1-1} \cmidrule{3-11}
PlenOctree-SH9 &  & 521.1 & 255.6 & 116.7 & 183.0 & 275.1 & 132.3 & 519.4 & 90.6 & 261.7 \\
PlenOctree-SH16 &  & 352.4 & 175.9 & 85.6 & 95.5 & 186.8 & 64.2 & 324.9 & 56.0 & 167.7 \\
PlenOctree-SH25 &  & 269.2 & 126.7 & 67.0 & 66.4 & 127.1 & 48.9 & 279.2 & 41.3 & 128.2 \\
PlenOctree-SG25 &  & 306.6 & 151.9 & 74.1 & 104.3 & 153.3 & 51.0 & 294.2 & 69.6 & 150.6 \\
\bottomrule
\end{tabular}
    \vspace{0.1em}
    \caption{\textbf{Per-scene quantitive results on spherical basis function ablations.}
    }
    \label{tab:per_scene_sh_sg}
\end{table*}

\end{document}